\documentclass[10pt]{article} %
\usepackage[accepted]{tmlr}

\usepackage{hyperref}
\usepackage{url}
\usepackage[utf8]{inputenc} %
\usepackage[T1]{fontenc}    %
\usepackage{xcolor}
\usepackage{algorithm}
\usepackage{amsfonts}       %
\usepackage{amsmath}
\usepackage{amssymb}
\usepackage{amsthm}
\usepackage{array}
\usepackage{bm}
\usepackage{booktabs}       %
\usepackage{caption}
\usepackage{colortbl}
\usepackage{empheq}
\usepackage{enumitem}
\usepackage{etoolbox}
\usepackage{float}
\usepackage{makecell}
\usepackage{mathrsfs}
\usepackage{mdframed}
\usepackage{microtype}      %
\usepackage{multirow}
\usepackage{multicol}
\usepackage{nccmath}
\usepackage{nicefrac}       %
\usepackage[noend]{algorithmic}
\usepackage{pifont}
\usepackage{ragged2e}
\usepackage{rotating}
\usepackage{subcaption}
\usepackage{tabularx}
\usepackage{tikz}
\usetikzlibrary{tikzmark}
\usepackage{url}            %
\usepackage{wrapfig}
\usepackage{color, colortbl}
\usepackage{xspace}
\usepackage{thm-restate}
\usepackage{xfrac}

\newcounter{rowcntr}[table]
\renewcommand{\therowcntr}{\arabic{chapter}.\the\numexpr\arabic{table}+1.\arabic{rowcntr}}

\AtBeginEnvironment{tabular}{\setcounter{rowcntr}{0}}
\newcolumntype{H}{>{\setbox0=\hbox\bgroup}c<{\egroup}@{}}

\newcommand{\PreserveBackslash}[1]{\let\temp=\\#1\let\\=\temp}
\newcolumntype{C}[1]{>{\centering\arraybackslash}m{#1}}
\newcolumntype{R}[1]{>{\raggedleft\arraybackslash}m{#1}}
\newcolumntype{L}[1]{>{\raggedright\arraybackslash}m{#1}}

\usepackage[capitalize]{cleveref}
\usepackage{pifont}
\crefname{section}{Sec.}{Secs.}
\Crefname{section}{Section}{Sections}
\Crefname{table}{Table}{Tables}
\crefname{table}{Tab.}{Tabs.}
\Crefname{appendix}{Appendix}{Appendices}
\crefname{appendix}{Appx.}{Apps.}

\newcommand{\comment}[1]{}

\def\VV{{\boldsymbol V}}
\def\UU{{\boldsymbol U}}

\definecolor{colorYes}{RGB}{51,160,44}
\definecolor{colorNo}{RGB}{228,26,28} %
\definecolor{darkgreen}{RGB}{34, 139, 34}
\newcommand{\cmark}{\textcolor{colorYes}{\ding{51}}}%
\newcommand{\xmark}{\textcolor{colorNo}{\ding{55}}}%

\definecolor{mintbg}{rgb}{.63,.79,.95}
\colorlet{lightmintbg}{mintbg!30}
\definecolor{pumpkin}{rgb}{0.75, 0.75, 0.75}
\colorlet{lightpumpkin}{pumpkin!20}

\def\imagenetval{{IN-val}}
\def\flickrval{{Flickr30k}}
\def\avgTE{{Avg. 38}}

\def\mobileclip{{MobileCLIP}}
\def\datacompdr{{DataCompDR}}
\def\datacompdrOneB{{DataCompDR-1B}}
\def\datacompdrTM{{DataCompDR-12M}}
\def\datacomp{{DataComp}}
\def\datacompOneB{{DataComp-1B}}
\def\datacompTM{{DataComp-1B12M}}
\def\dfn{{DFN}}
\def\dfndr{{DFNDR}}
\def\dfnFiveB{{DFN-5B}}
\def\dfnTwoB{{DFN-2B}}
\def\dfnFiveTM{{DFN-5B12M}}
\def\dfnTwoTM{{DFN-2B12M}}
\def\dfnTM{{DFN-12M}}

\def\dfndrTwoB{{DFNDR-2B}}
\def\dfndrFiveTM{{DFNDR-5B12M}}
\def\dfndrTwoTM{{DFNDR-2B12M}}
\def\dfndrTM{{DFNDR-12M}}
\def\mci{{MCi}}
\def\mciZero{{MCi0}}
\def\mciOne{{MCi1}}
\def\mciTwo{{MCi2}}
\def\mciThree{{MCi3}}
\def\mciFour{{MCi4}}
\def\mct{{MCt}}
\def\mobileclipB{{MobileCLIP-B}}
\def\mobileclipSTwo{{MobileCLIP-S2}} %
\def\mobileclipSOne{{MobileCLIP-S1}} %
\def\mobileclipSZero{{MobileCLIP-S0}} %
\def\mobileclipL{MobileCLIP-S3}
\def\mobileclipXL{MobileCLIP-S4}
\def\mobileclipLFourteen{MobileCLIP-L/14}

\def\ourmodel{{MobileCLIP2}}
\def\ourmodelB{\ourmodel{}-B}
\def\ourmodelSTwo{\ourmodel{}-S2}

\def\ourmodelSZero{\ourmodel{}-S0}
\def\ourmodelL{\ourmodel{}-S3}
\def\ourmodelXL{\ourmodel{}-S4}
\def\ourmodelLFourteen{\ourmodel{}-L/14}

\newcommand{\LTotal}{{\mathcal{L}_{\text{Total}}}}
\newcommand{\LKD}{{\mathcal{L}_{\text{Distill}}}}

\newcommand{\LClip}{{\mathcal{L}_{\text{CLIP}}}}
\newcommand{\Similarity}{{\mathcal{S}}}

\newcommand{\FeatStudentImg}{{\varPhi_{\text{img}}}}
\newcommand{\FeatStudentTxt}{{\varPhi_{\text{txt}}}}

\newcommand{\FeatTeacherImgk}{{\Psi_{\text{img}}^{(k)}}}
\newcommand{\FeatTeacherTxtk}{{\Psi_{\text{txt}}^{(k)}}}

\usepackage[textwidth=2.5cm]{todonotes}

\title{\ourmodel{}: Improving Multi-Modal Reinforced Training}

\author{
    \name Fartash Faghri\thanks{Equal contribution.} \email fartash@apple.com\\
    Apple
    \AND
    \name Pavan Kumar Anasosalu Vasu$^*$ \email panasosaluvasu@apple.com\\
    Apple
    \AND
    \name Cem Koc \email cem\_koc@apple.com\\
    Apple
    \AND
    \name Vaishaal Shankar \email \\
    Work done while at Apple
    \AND
    \name Alexander Toshev \email toshev@apple.com\\
    Apple
    \AND
    \name Oncel Tuzel \email otuzel@apple.com\\
    Apple
    \AND
    \name Hadi Pouransari \email mpouransari@apple.com\\
    Apple
}

\def\month{08}  %
\def\year{2025} %
\def\openreview{\url{https://openreview.net/forum?id=WeF9zolng8}} %

\begin{document}

\maketitle

\begin{abstract}
    Foundation image-text models such as CLIP with zero-shot capabilities 
    enable a wide array of applications. MobileCLIP is a recent family of 
    image-text models at 3--15ms latency and 50--150M parameters with 
    state-of-the-art zero-shot accuracy. The main ingredients in MobileCLIP 
    were its low-latency and light architectures and a novel multi-modal 
    reinforced training that made knowledge distillation from multiple 
    caption-generators and CLIP teachers efficient, scalable, and reproducible.  
    In this paper, we improve the multi-modal reinforced training of MobileCLIP 
    through:
    1) better CLIP teacher ensembles trained on the DFN dataset,
    2) improved captioner teachers trained on the DFN dataset and fine-tuned on 
       a diverse selection of high-quality image-caption datasets.
    We discover new insights through ablations such as
    the importance of temperature tuning in contrastive knowledge distillation,
    the effectiveness of caption-generator fine-tuning for caption diversity, 
    and the additive improvement from combining synthetic captions generated by 
    multiple models.
    We train a new family of models called \ourmodel{} and achieve state-of-the-art ImageNet-1k zero-shot accuracies at low latencies. In particular, 
    we observe 2.2\% improvement in ImageNet-1k accuracy for \ourmodelB{} compared with MobileCLIP-B 
    architecture.
    Notably, \ourmodelXL{} matches the zero-shot accuracy of SigLIP-SO400M/14 on ImageNet-1k while being 2$\times$ smaller and improves on DFN ViT-L/14 at 2.5$\times$ lower latency.
    We release our pretrained models~\footnote{\url{https://github.com/apple/ml-mobileclip}} and the data generation code~\footnote{\url{https://github.com/apple/ml-mobileclip-dr}}. The 
    data generation code makes it easy to create new reinforced datasets with 
    arbitrary teachers using distributed scalable processing.
\end{abstract}

\section{Introduction}

CLIP~\citep{CLIP} is an image-text model that maps images and text inputs to 
a shared embedding space, where a text describing an image, also called 
caption, is mapped close to an image matching its description but far from 
dissimilar images. Building on a vast
literature~\citep{frome2013devise,socher2014grounded,karpathy2015deep,kiros2014unifying,faghri2018vse++},
CLIP substantially increased the scale of training data and models.  
Consequentially, along with improved image-text retrieval performance, new 
zero-shot classification capabilities emerged with non-trivial accuracy on 
classification tasks without any explicit supervised training with 
classification labels through linear probing. The image-encoder can be further 
specialized to a new task by either linear probing (fixed encoder), or full 
fine-tuning to achieve state-of-the-art performance on a diverse set of 
tasks~\citep{wortsman2022robust}. CLIP is one of the first to be called 
a foundation model given the diversity of its capabilities and 
applications~\citep{bommasani2021opportunities}.

\begin{wrapfigure}{r}{0.6\linewidth}
\vspace{-10pt}
    \centering
    \includegraphics[width=\linewidth]{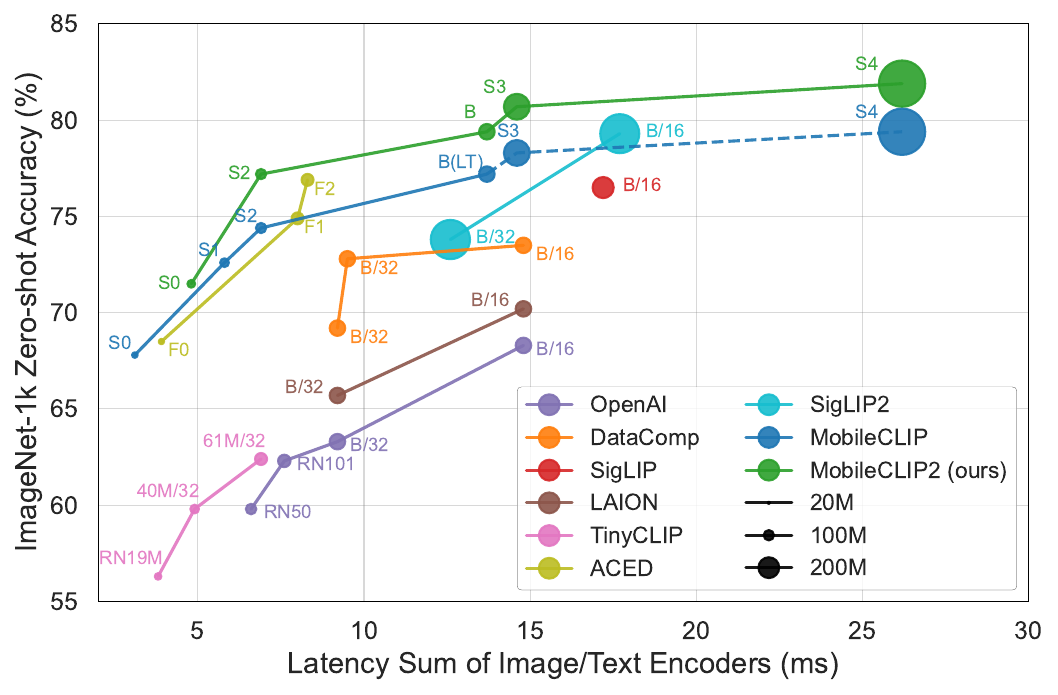}
    \vspace{-21pt}
    \caption{\textbf{\ourmodel{} models trained on
    \dfndrTwoB{} achieve state-of-the-art accuracy at 
    low latencies.}
    \ourmodelXL{} matches the accuracy of SigLIP-SO400M/14 with 2$\times$ fewer parameters and
    surpasses DFN ViT-L/14 at 2.5$\times$ lower latency measured on iPhone12 Pro Max.
    \mobileclip{}-S3/S4 are our new architectures trained
    on MobileCLIP's training dataset, \datacompdrOneB{} (dashed lines).
    }\label{fig:accuracy_vs_latency_tradeoff}
\end{wrapfigure}

The success of CLIP resulted in an increase in the sizes of models and datasets, leading to 
a gradual increase in performance~\citep{EVA02,SigLIP,gadre2024datacomp,fang2023data}.  Recently, 
this trend has been reversed to models with small size and low latency for 
applications on mobile devices.
Notably, TinyCLIP~\citep{tinyclip} and MobileCLIP~\citep{vasu2024mobileclip} 
proposed models with as few as 50M total parameters (sum of image and text 
encoder parameters). For example, MobileCLIP-S0, with total latency of 3ms (sum 
of image and text encoder latencies), achieves similar average performance to 
the original OpenAI ViT-B/16 CLIP while being 3x smaller and 5x faster.  It 
also demonstrates improved performance compared to prior state-of-the-art 
larger models, such as SigLIP~\citep{SigLIP}.

In this paper, we present ablations of multi-modal reinforced training and 
present an improved training recipe. We train a new family of models, 
\ourmodel{}, that establishes new state-of-the-art ImageNet-1k accuracy at a range of latencies
matching the performance of larger SigLIP~\citep{SigLIP} and DFN~\citep{fang2023data} models while up to 4$\times$ smaller (our \ourmodelSTwo{} compared with SigLIP2-B/32) and up to 2.5$\times$ faster (our \ourmodelXL{} compared with DFN ViT-L/14). Moreover, we release 
efficient distributed code for generating reinforced datasets using arbitrary 
teacher models.

\section{Improved Training}
\label{sec:method}
\mobileclip{} introduced a family of low-latency image-text models consisting 
of S0, S1, S2, B, and B-LT variants with aggregate image-text latencies spanning
{3.8}-{13.7}ms. These low latencies were achieved with specialized 
architectures based on FastViT~\citep{vasu2023fastvit} and an improved training 
method called multi-modal reinforced training. We seek to further explore and 
improve each step of multi-modal reinforced training. We additionally consider 
a more diverse family of architectures that cover a wider range of latencies.

Reinforced training is a method for achieving better performance from a base 
dataset through improvements from additional sources such as pretrained 
models~\citep{faghri2023reinforce}. Multi-modal reinforced training introduced 
in \citet{vasu2024mobileclip} adds information to an image-text dataset from 
pretrained image-text models as well as a pretrained synthetic caption 
generator.  Specifically, they add the following additional information to 
\datacompOneB{} dataset: 1) image embeddings from two CLIP teachers on 10 
random augmentations of each image 2) text embeddings from two CLIP teachers on 
the original text as well as 5 synthetic captions generated from a CoCa caption 
generator. Given a reinforced dataset, they modify the training loss to include 
a knowledge distillation loss given the embeddings from teachers on each 
sample~\citep{hinton2015distilling}. To ensure consistency between the teacher 
and student, the same image augmentation is reproduced via stored augmentation 
parameters~\citep{beyer2022knowledge,faghri2023reinforce}. They perform 
ablations to find the set of CLIP teachers, caption generator, and image 
augmentations that provide the largest performance gain on ImageNet as well as 
the average accuracy on 38 evaluations from DataComp~\citep{gadre2024datacomp}.

We follow a similar multi-modal reinforced training to \mobileclip{} while 
improving all aspects and call the resulting model family \ourmodel{}.  
\Cref{tab:method_summary} summarizes the gains from each major improvement. In 
short, we use better training data, better CLIP teacher models, and better and 
diverse synthetic caption generators compared to \mobileclip{}. In all 
ablations, we train \mobileclipB{} for 30k iterations ($\sim$20 epochs) on 
datasets with 12.8M images. We provide a summary of datasets in this paper in \cref{tab:dataset_summary}.

\begin{figure}[t]
    \centering
    \includegraphics[width=0.49\linewidth]{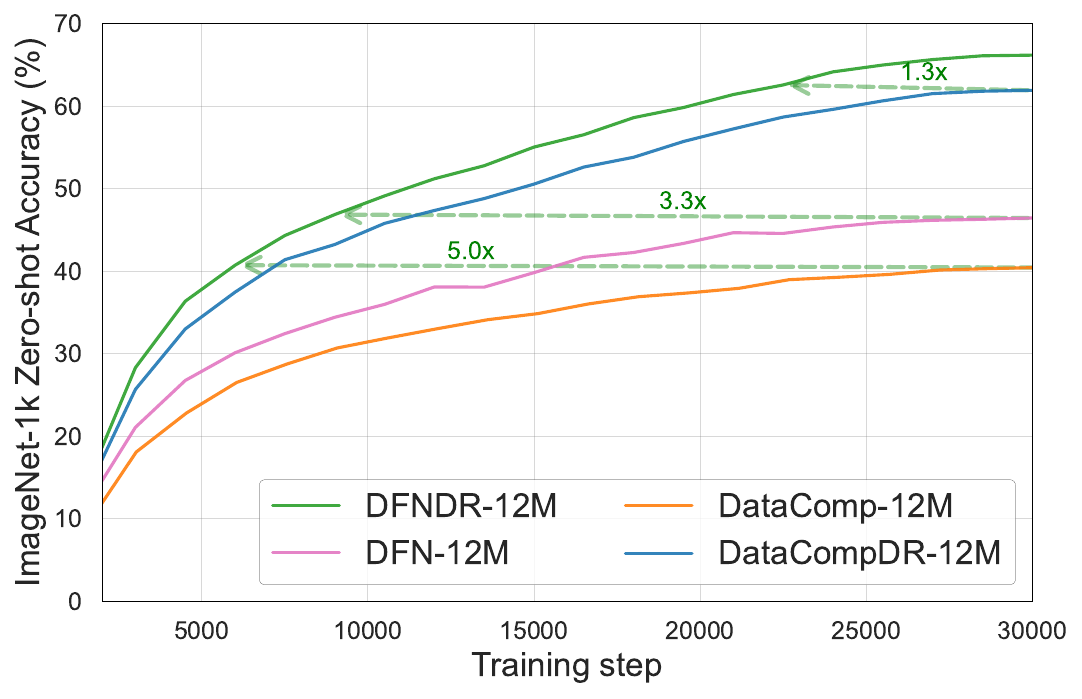}
    \hfill
    \includegraphics[width=0.49\linewidth]{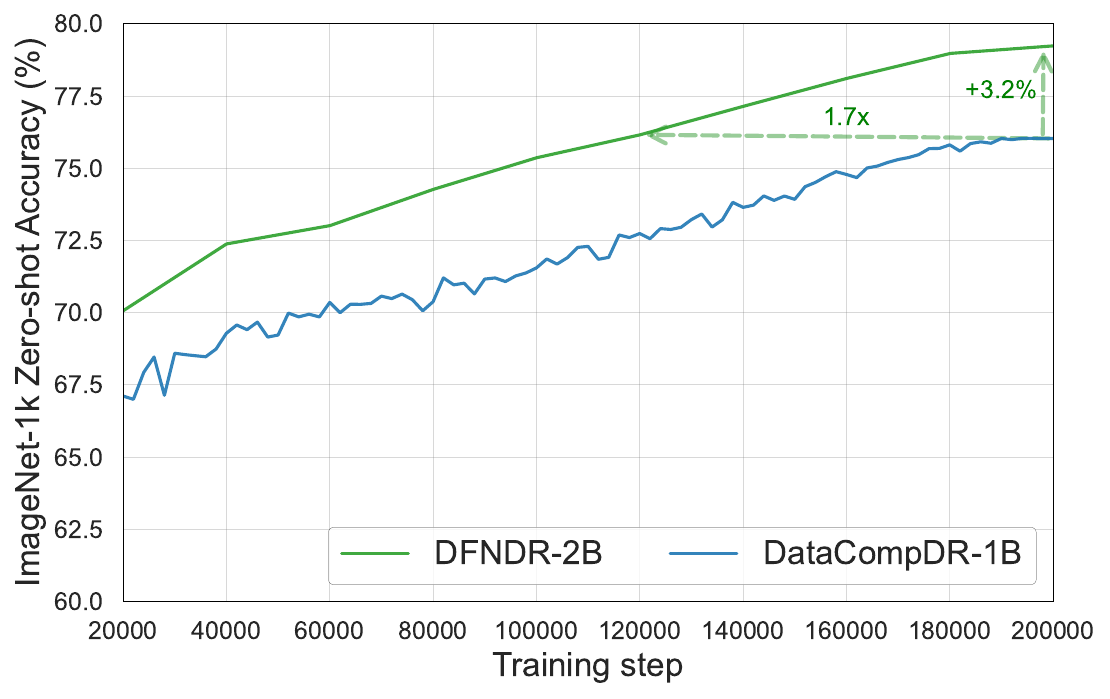}
    \vspace{-2pt}
    \caption{Left: \textbf{Training on \dfndrTM{} is up to 5x more 
    efficient} compared with \datacompTM{}, 3.3x compared with \dfnTM{}, and 
    {1.3x} compared to \datacompdrTM{}.  All models are trained for 30k iterations and global batch size 8192 (246M seen samples). \dfnTM{} consists of 12M uniformly sampled image-text pairs from \dfnTwoB{} and \dfndrTM{} consists of additional reinforcements per sample in \dfnTM{}. Right: \textbf{Training on 
    \dfndrTwoB{} is up to {1.7x} more efficient} compared with 
    \datacompdrOneB{}. Models are trained for 200k iterations and gloabl batch size 65536 (13B seen samples).}
    \label{fig:training_efficiency}
\end{figure}

\Cref{fig:training_efficiency} demonstrates the efficiency gains compared with 
\dfn{}~\citep{fang2023data}, \datacomp{}~\citep{gadre2024datacomp} and 
\datacompdr{}~\citep{vasu2024mobileclip} datasets during training.  Training on 
\dfndrTwoTM{} for 30M seen samples is 5x more efficient than training on 
\datacompTM{}, i.e., we reach the ImageNet-1k zero-shot accuracy of training on 
\datacompTM{} for 30M after seeing only 6M samples of \dfndrTwoTM{}.  
Similarly, training on \datacompdrTM{} is 3.3x more efficient compared to 
\dfnTwoTM{} and 1.3x more efficient compared with \datacompdrTM{}.  We also 
observe {1.6x} speedup when training on \dfndrTwoB{} compared with training on 
\datacompdrOneB{} for 13B seen samples.
Similar to \datacompdr{}, training on \dfndr{} datasets do not have any 
wall-clock time overhead, i.e., each training step of training on \datacomp{}, 
\dfn{}, \datacompdr{}, and \dfndr{} takes the same amount of time.  That means 
any efficiency gains in terms of the number of samples and training iterations 
directly translate to wall-clock time efficiency gains.

\begin{table}[t]
    \centering
    \caption{\textbf{Summary of \ourmodel{} training improvements.}
    \label{tab:method_summary}
    CoCa models are pretrained on a large dataset for 13B seen samples then 
    fine-tuned for 12M seen samples (denoted by $\rightarrow$).
    The architecture for all CLIP teachers in this table is ViT-L/14.
    We report mean and standard deviations of 5 runs when available.
    }
        \resizebox{0.95\columnwidth}{!}{
    \begin{tabular}{lccc|ccc}
        \toprule[1.5pt]
Name & Dataset & CLIP Teacher Datasets & CoCa Dataset & \imagenetval{}& \flickrval{}& \avgTE{} \\ %
\midrule[1.25pt]
\mobileclip~\citep{vasu2024mobileclip}
& \datacompTM{} & OpenAI + DataComp-XL & LAION-2B $\rightarrow$ MSCOCO-123k
& 61.6 & 72.8 & 53.5 \\ %
\Cref{tab:datacomp_vs_dfn}
& \cellcolor{lightmintbg}\dfnFiveTM{}  & OpenAI + DataComp-XL & LAION-2B $\rightarrow$ MSCOCO-123k  
& 63.1$_{0.2}$ & 73.3$_{0.6}$ & 54.1$_{0.4}$ \\ %
\Cref{tab:clip_teachers_ensemble}
& \dfnFiveTM{}      & \cellcolor{lightmintbg}DFN-2B + DFN-2B-s39B & LAION-2B $\rightarrow$ MSCOCO-123k
& 65.4$_{0.4}$ & \cellcolor{lightmintbg}75.8$_{0.3}$ & \cellcolor{lightpumpkin}56.2$_{0.6}$ \\ %
\ourmodel{} (\cref{tab:capgen_ft})
&  \dfnFiveTM{} & DFN-2B + DFN-2B-s39B & \cellcolor{lightmintbg} \dfnTwoB{} $\rightarrow$ MSCOCO-38k
& \cellcolor{lightmintbg}65.9$_{0.3}$ & \cellcolor{lightpumpkin}75.4$_{0.2}$ & \cellcolor{lightmintbg}56.5$_{0.3}$ \\ %
\Cref{tab:capgen_ft}
& \dfnFiveTM{}  & DFN-2B + DFN-2B-s39B & \cellcolor{lightmintbg} \dfnTwoB{} $\rightarrow$ Syn.$\times$10
& \cellcolor{lightmintbg}66.0$_{0.1}$ & 75.1$_{0.6}$ & \cellcolor{lightmintbg}56.5$_{0.3}$ \\ %
        \bottomrule[1.5pt]
    \end{tabular}}

\end{table}

\subsection{Multi-Modal Reinforced Training}

Dataset Reinforcement (DR)~\citep{faghri2023reinforce} is a method for 
improving a dataset to achieve higher accuracy with minimal changes to the 
training code and minimal computational overhead. DR was first introduced for 
training image classifiers where \citet{faghri2023reinforce} improved the 
ImageNet dataset by storing classification probabilities efficiently from 
a strong ensemble of classifiers. Given stored probabilities, the training was 
essentially Knowledge Distillation~\citep{hinton2015distilling} with no 
overhead for computing the teacher predictions. The cost efficiency makes it 
feasible to train longer for larger gains as observed in 
\citet{beyer2022knowledge}.
\citet{vasu2024mobileclip} adopted DR for training image-text CLIP models by 
storing knowledge from a strong ensemble of CLIP models and additionally 
synthetic captions from an image caption generator. They demonstrated 
up to 1000$\times$ improved learning efficiency compared with non-reinforced 
CLIP training.

Given a batch of $b$ image-text pairs, we denote the embeddings of a target student model by
$\FeatStudentImg, \FeatStudentTxt \in \mathcal{R}^{b \times d}$, where $d$ is the dimensionality
of the shared embedding space. We utilize two types of teachers, an image-text teacher ensemble that maps images and texts to a shared space similar to CLIP~\citep{CLIP} and CoCa-based captioners that generate
a caption given an image using an encoder-decoder architecture~\citep{CoCa}.
Let $\FeatTeacherImgk, \FeatTeacherTxtk \in \mathcal{R}^{b \times d_k}$ denote
the image-text embeddings from the $k$-th CLIP-based teacher where $d_k$ is the dimensionality
of the shared space. The distillation loss is defined as
\begin{align}
\LKD &= \frac{1}{2bK} \sum_{k=1}^{K}
\underbrace{\text{KL}(
    \Similarity_{\tau_k}(\FeatTeacherImgk, \FeatTeacherTxtk)
    \|
    \Similarity_{\widehat{\tau}}(\FeatStudentImg, \FeatStudentTxt))}
    _{\text{Image to Text}}
+
\underbrace{\text{KL}(
    \Similarity_{\tau_k}(\FeatTeacherTxtk, \FeatTeacherImgk)
    \|
    \Similarity_{\widehat{\tau}}(\FeatStudentTxt, \FeatStudentImg))}
    _{\text{Text to Image}},
    \label{eq:kl_loss}
\end{align}
where KL denotes Kullback-Leibler divergence, and $\Similarity_\tau(\UU, \VV)$ is the row-wise
Softmax operation applied to $\UU\VV^{\top}/\tau$ with temperature $\tau$. The total loss is
    $\LTotal = (1-\lambda) \LClip + \lambda \LKD\,$,
that is the sum of a standard CLIP loss and the distillation loss with coefficients $1-\lambda$ and $\lambda$, respectively.

\subsection{Better Base Dataset: DFN}
Multi-modal reinforced training starts from a base dataset containing real 
image-text pairs commonly collected from the web.  
DataComp~\citep{gadre2024datacomp} demonstrated that the quality of large-scale 
image-text datasets can be significantly improved through filtering based on 
scores such as compatibility of image and text. Their proposed BestPool 
filtering applied on a pool of 12B samples resulted in the \datacompOneB{} 
dataset that was used as the base dataset in \mobileclip{}. DataComp also 
released the original 12B samples as a benchmark for dataset curation and 
filtering methods. DFN~\citep{fang2023data} proposed to filter data using 
a filtering network trained on high-quality data. Applying their model on 
DataComp-12B pool resulted the \dfnTwoB{} dataset. They additionally collected 
a larger set of images from the web disjoint from DataComp-12B and after 
filtering resulted in another 3B samples and collectively created the 
\dfnFiveB{} dataset.

We study the impact of replacing the base dataset in \mobileclip{} with 
\dfnFiveB{}. We ablate using 12M uniformly sampled subset of \datacompOneB{} 
referred to as \datacompTM{} that was introduced in \citep{vasu2024mobileclip} 
for rapid experimentation. We similarly sample a 12M subset from \dfnFiveB{} 
referred to as \dfnFiveTM{}.  \Cref{tab:datacomp_vs_dfn} compares the 
performance of training with and without distillation/synthetic captions. We 
observe that \dfnFiveTM{} results in up to 1.4\% gain together with 
distillation and synthetic captions. Although this gain is smaller compared to 
the up to 6\% gain without distillation/synthetic captions, it is still more 
than standard deviation.

\begin{table}[h]
    \centering
    \caption{\textbf{Training on DFN is better than DataComp with and without distillation/synthetic captions.}
    CLIP teachers and synthetic caption generators are the same as 
    \mobileclip{} (OpenAI+DataComp-XL CLIP-ViT-L/14, and CoCa-ViT-L/14).
    For distillation, the coefficient $\lambda$ is set to $1.0$ (no CLIP loss) and use strong image augmentations.
    }
    \label{tab:datacomp_vs_dfn}
        \resizebox{0.55\columnwidth}{!}{
    \begin{tabular}{ccc|ccc}
        \toprule[1.5pt]
        Dataset       & Distill. & Syn. Caps. & \imagenetval{}& \flickrval{}& \avgTE{}\\
        \midrule[1.25pt]
        \datacompTM{} & \xmark   & \xmark     & 44.6 & 42.4 & 40.1 \\
        \dfnFiveTM{}      & \xmark   & \xmark     & 49.9 & 48.5 & 43.5 \\
        \midrule[1pt]
        \datacompTM{} & \xmark   & \cmark     & 51.9 & 71.8 & 47.8 \\
        \dfnFiveTM{}      & \xmark   & \cmark     & 54.9 & \cellcolor{lightpumpkin}70.7 & 49.6 \\
        \midrule[1pt]
        \datacompTM{} & \cmark   & \xmark     & 56.3 & 57.8 & 48.7 \\
        \dfnFiveTM{}      & \cmark   & \xmark     & 59.5 & 60.4 & 50.0 \\
        \midrule[1pt]
        \datacompTM{} & \cmark   & \cmark     & 61.6 & 72.8 & 53.7 \\
        \dfnFiveTM{}      & \cmark   & \cmark     & 63.0 & 74.1 & 54.6 \\
        \bottomrule[1.5pt]
    \end{tabular}}

\end{table}

\subsection{DFN CLIP Teachers}
\label{sec:method_clip_teachers}
One source of reinforcement in multi-modal reinforced training is the 
embeddings from CLIP teachers that are used as targets in CLIP distillation.  
\citep{vasu2024mobileclip} performed a comprehensive study of existing strong 
CLIP teachers at the time of publication and found the ensemble of 
\texttt{ViT-L-14-openai} and \texttt{ViT-L-14-datacomp\_xl\_s13b\_b90k} to 
result the best student performance. Here we investigate the effectiveness of 
DFN-pretrained models as teachers. DFN-pretrained CLIP models with ViT-L-14 and 
ViT-H-14 achieve the state-of-the-art performance on Avg. 38 evaluations of 
DataComp~\citep{fang2023data} better than other popular models such as 
SigLIP~\citep{SigLIP}.

As the choice of the caption generator and the CLIP teachers may depend on each 
other, we reduce the complexity of our analysis by analyzing the effect of the 
CLIP teachers on synthetic captions from a CoCa model without fine-tuning (See \cref{sec:method_capgen}).  We 
explore the diversity of synthetic captions through fine-tuning in 
\cref{sec:method_capgen_ft}.

\begin{table}[h]
    \centering
    \caption{\textbf{Optimal logit scale for distillation varies across teachers.}
    The dataset is \dfnFiveTM{} with synthetic captions generated from CoCa-DFN-2B in \cref{sec:method_capgen}.  The loss coefficient $\lambda$ is set to 
    $1.0$ and trained using strong image augmentations.}
    \label{tab:clip_teachers_single}
        \resizebox{0.7\columnwidth}{!}{
    \begin{tabular}{cc|ccc}
        \toprule[1.5pt]
        Teacher & Logit Scale & \imagenetval{}& \flickrval{}& \avgTE{}\\
        \midrule[1.25pt]
\texttt{datacomp\_xl\_s13b\_b90k-CLIP-ViT-L-14} & 50 & 62.6 & 65.6 & 53.3\\
\texttt{DFN2B-CLIP-ViT-L-14}                 & 70 & 65.5 & 68.0 & 56.5\\
\texttt{DFN5B-CLIP-ViT-H-14}                 & 90 & 64.0 & 65.9 & 54.7\\
\texttt{DFN5B-CLIP-ViT-H-14-384}             & 55 & 64.6 & 67.6 & 54.4\\
\texttt{DFN2B-CLIP-ViT-L-14-s39b}            & 60 & 65.2 & 67.5 & 54.8\\
        \bottomrule[1.5pt]
    \end{tabular}}

\end{table}

\begin{table}[h]
    \centering
    \caption{\textbf{Ensemble of DFN CLIP teachers improve ImageNet-1k validation accuracy by 2.8\%.}
    The dataset is \dfnFiveTM{} with synthetic captions generated from CoCa-DFN-2B in \cref{sec:method_capgen}.  The loss coefficient $\lambda$ is set to 
    $1.0$ and trained using strong image augmentations. The optimal logit scales for each model is set independently based on \cref{tab:clip_teachers_single}.}
    \label{tab:clip_teachers_ensemble}
        \resizebox{0.8\columnwidth}{!}{
    \begin{tabular}{cc|ccc}
        \toprule[1.5pt]
        Teacher 1 & Teacher 2 & \imagenetval{}& \flickrval{}& \avgTE{}\\
        \midrule[1.25pt]
\texttt{ViT-L-14-openai}
& \texttt{ViT-L-14-datacomp\_xl\_s13b\_b90k}
& 63.1 & 64.7 & 55.2\\
        \midrule

\texttt{ViT-L-14-datacomp\_xl\_s13b\_b90k}
& \texttt{DFN5B-CLIP-ViT-H-14-384}
& 64.5 & 67.8 & 54.5\\

\texttt{ViT-L-14-datacomp\_xl\_s13b\_b90k}
& \texttt{DFN5B-CLIP-ViT-H-14}
& 64.4 & 67.3 & 55.3\\

\texttt{ViT-L-14-datacomp\_xl\_s13b\_b90k}
& \texttt{DFN2B-CLIP-ViT-L-14}
& 65.3 & 68.1 & \cellcolor{lightmintbg} 56.2\\

\texttt{DFN5B-CLIP-ViT-H-14-384}
& \texttt{DFN5B-CLIP-ViT-H-14}
& 64.7 & 66.9 & 54.9\\

\texttt{DFN5B-CLIP-ViT-H-14-384}
& \texttt{DFN2B-CLIP-ViT-L-14}
& \cellcolor{lightmintbg} 65.8 & \cellcolor{lightmintbg} 68.6 & \cellcolor{lightmintbg} 56.2\\

\texttt{DFN5B-CLIP-ViT-H-14}
& \texttt{DFN2B-CLIP-ViT-L-14}
& 65.2 & 68.0 & 55.8\\

\texttt{DFN2B-CLIP-ViT-L-14-s39b}
& \texttt{datacomp\_xl\_s13b\_b90k}
& 65.1 & 67.6 & 55.7\\

\texttt{DFN2B-CLIP-ViT-L-14-s39b}
& \texttt{DFN5B-CLIP-ViT-H-14-384}
& \cellcolor{lightmintbg} 65.7 & 67.3 & 55.1\\

\texttt{DFN2B-CLIP-ViT-L-14-s39b}
& \texttt{DFN5B-CLIP-ViT-H-14}
& \cellcolor{lightmintbg} 65.7 & 68.2 & 55.7\\

\texttt{DFN2B-CLIP-ViT-L-14-s39b}
& \texttt{DFN2B-CLIP-ViT-L-14}
& \cellcolor{lightmintbg} 65.9 & \cellcolor{lightmintbg} 68.7 & \cellcolor{lightmintbg} 55.9\\
        \bottomrule[1.5pt]
    \end{tabular}}

\end{table}

\paragraph{Logit scaling.} CLIP models are trained with a logit scale that is 
tuned during the training in the range 0-100. MobileCLIP used the same logit 
scalar as the temperature scaling in the KD loss. We observe that the logit 
scalar in DFN and DataComp models is not optimal for KD and tune that further.  
\Cref{tab:clip_teachers_single} shows the optimal logit scale used for each 
teacher to train a \mobileclipB{} model. We observe that the logit scale is not a sensitive hyperparameter where values within a range of 5 points achieve similar performance.

\paragraph{Ensemble teachers.} We construct ensembles of size two using 
DataComp and DFN teachers. \Cref{tab:clip_teachers_ensemble} shows the 
performance of training a \mobileclipB{} model using embeddings from various 
ensembles. We observe significant improvements compared with teachers used in 
{MobileCLIP}. Specifically, \imagenetval{} and \flickrval{} improve by up to 
3\%.  We choose the ensemble of \texttt{DFN2B-CLIP-ViT-L-14-s39b} and 
\texttt{DFN2B-CLIP-ViT-L-14} for \ourmodel{} based on its performance and cost 
efficiency compared to other larger or higher resolution ensembles.
We utilize the optimal logit scales for each member of the ensemble that is found independently.
It is possible that the optimal logit scales for ensemble would vary when used together but we do not further optimize logit scales jointly.

\subsection{DFN Caption Generators}
\label{sec:method_capgen}
Another source of reinforcements for training \ourmodel{} is synthetic captions 
generated from an image caption generator. \mobileclip{} used a single CoCa 
captioner which has a two-tower image-text architecture coupled with a text 
decoder~\citep{CoCa}. Compared with most recent VLMs, the text-decoder is 
fairly light-weight that results in an overall relatively faster caption 
generator compared with more recent VLMs~\citep{liu2024visual,vasu2024fastvlm}. As 
\mobileclip{} generated multiple synthetic captions on billions of images, the 
cost of running CoCa was an important decision factor.
They did not provide analysis on the choice of captioner but observed 
significant gains from training on synthetic captions compared with not using 
synthetic captions (7.4\% for 30k training iterations).  \mobileclip{} 
generated {5} synthetic captions per image although they observed the majority 
of the gain comes from the first 1-2 synthetic captions.

We explore training a new CoCa model using the DFN dataset to 
improve the quality of synthetic captions. We adopt the same architecture as 
the CoCa model utilized in \mobileclip{} based on the ViT-L/14 image encoder.  
They utilized the model trained on LAION-2B dataset and fine-tuned on 
MSCOCO-128k dataset. We pretrain the same architecture on \dfnTwoB{} for 13B 
seen samples using OpenCLIP~\citep{OpenCLIP}.

\begin{table}[h]
    \centering
    \caption{\textbf{Pretraining CoCa on \dfnTwoB{} without fine-tuning results 
    in similar IN-1k performance but worse robustness and retrieval.}
    The dataset is \dfnFiveTM{}, CLIP teachers are the same as \mobileclip{} 
    (OpenAI+DataComp-XL CLIP-ViT-L/14) and the architecture of the CoCa model 
    is the same as CoCa-ViT-L/14.  For distillation, the coefficient $\lambda$ is set to 
    $1.0$ (no CLIP loss) and use strong image augmentations. Values 
    within one standard deviation of the best of each group are highlighted.}
    \label{tab:coca_dfn}
        \resizebox{0.6\columnwidth}{!}{
    \begin{tabular}{cccc|ccc}
        \toprule[1.5pt]
        \multirow{2}{*}{Distill.} & \multirow{2}{*}{High Aug.}
        & \multicolumn{2}{c}{CoCa}
        & \imagenetval{}& \flickrval{}& \avgTE{}\\
        \cmidrule(lr){3-4}
        & & LAION-2B $\rightarrow$ MSCOCO-128k & DFN-2B \\
        \midrule[1.25pt]
\xmark & \xmark & \xmark & \xmark & 49.9 & 48.5 & 43.5 \\
\xmark & \xmark & \cmark & \xmark & 54.9 & \cellcolor{lightmintbg} 70.6 & \cellcolor{lightmintbg} 49.6 \\
\xmark & \cmark & \cmark & \xmark & 51.1 & 65.7 & 45.3 \\
\xmark & \cmark & \xmark & \cmark & 54.6 & 55.1 & 46.2 \\
\xmark & \cmark & \cmark & \cmark & \cellcolor{lightmintbg}56.8 & 67.2 & 48.4 \\
\midrule
\cmark & \cmark & \xmark & \xmark & 59.5 & 60.3 & 50.0 \\
\cmark & \cmark & \cmark & \xmark & 63.0 & \cellcolor{lightmintbg} 74.1 & 54.6 \\
\cmark & \cmark & \xmark & \cmark & 63.1 & 64.7 & \cellcolor{lightmintbg} 55.2 \\
\cmark & \cmark & \cmark & \cmark & \cellcolor{lightmintbg} 63.4 & 72.0 & \cellcolor{lightmintbg}55.1 \\
        \bottomrule[1.5pt]
    \end{tabular}}

\end{table}

\Cref{tab:coca_dfn} demonstrates the impact of DFN-CoCa synthetic captions on 
the performance with and without distillation. We observe that utilizing 
DFN-CoCa synthetic captions results in improved \imagenetval{} and \avgTE{} 
performance but negatively impacts retrieval. As we observe in 
\cref{sec:method_capgen_ft}, the retrieval performance recovers with 
fine-tuning on high-quality datasets such as {MSCOCO}. We further observe the 
synthetic captions from the original CoCa model can be used together with 
DFN-CoCa captions to provide additional gains but these gains are small with 
distillation.

\subsection{Fine-tuning Caption Generators}
\label{sec:method_capgen_ft}
In \cref{sec:method_capgen}, we showed that pretraining a CoCa model on 
\dfnTwoB{} results in improved \imagenetval{} and \avgTE{} performance when 
utilized for multi-modal reinforced training. However, the retrieval 
performance falls behind which is due to the lack of fine-tuning on 
a high-quality dataset. MobileCLIP used a CoCa model fine-tuned on 
MSCOCO~\citep{MSCOCO}. MSCOCO-2017 contains 123k images with captions that have higher quality compared to average 
image-text pairs in DataComp and DFN datasets.

In this section, we study the impact of fine-tuning on various high-quality 
datasets. In addition to 123k samples from MSCOCO which we refer to as 
MSCOCO-123k, we also consider a subset of 38k samples with permissive licenses 
(CC Attribution 2.0, CC Attribution-ShareAlike 2.0, and CC Attribution-NoDerivs 
{2.0}) which we refer to as MSCOCO-38k. We also consider 
GBC-1M/10M~\citep{hsieh2024graph}, 
DOCCI-9k-short/extended/complete~\citep{onoe2025docci}, 
DCI-8k~\citep{urbanek2024picture}, and ReCap-COCO-30k~\citep{li2024if}. We 
fine-tune DFN-CoCa on each dataset for 12M seen samples using the same loss as 
CoCa pretraining.

\begin{table}[h]
    \centering
    \caption{\textbf{}
    The dataset is \dfnFiveTM{}, CLIP teachers are our selected DFN models 
    (DFN2B-CLIP-ViT-L-14-s39b and DFN2B-CLIP-ViT-L-14) and the architecture of the CoCa model 
    is the same as CoCa-ViT-L/14.  For distillation, the coefficient is set to 
    $1.0$ (no CLIP loss) and use strong image augmentations.  
    }
    \label{tab:capgen_ft}
        \resizebox{0.6\columnwidth}{!}{
    \begin{tabular}{ccc|ccc}
        \toprule[1.5pt]
        Base Dataset & FT Dataset & Context len. & \imagenetval{}& \flickrval{}& \avgTE{}\\
        \midrule[1.25pt]
LAION-2B & MSCOCO-123k & 77 & 65.4$_{0.4}$ & \cellcolor{lightmintbg} 75.8$_{0.3}$ & 56.2$_{0.6}$\\
\midrule
DFN-2B & - & 77 & \cellcolor{lightmintbg} 65.9 & 68.7 & 55.9\\
DFN-2B & MSCOCO-123k & 77 & \cellcolor{lightmintbg} 65.9 & \cellcolor{lightmintbg} 76.0 & 56.2\\
DFN-2B & MSCOCO-38k & 77 & \cellcolor{lightmintbg} 65.9$_{0.3}$ & \cellcolor{lightpumpkin} 75.4$_{0.2}$ & \cellcolor{lightpumpkin}56.5$_{0.3}$ \\
DFN-2B & GBC1M-short & 77 & 65.8 & 75.0 & 56.6\\
DFN-2B & DOCCI & 77 & 66.3 & 72.6 & \cellcolor{lightmintbg} 57.3\\
DFN-2B & DCI-short & 77 & \cellcolor{lightmintbg} 65.9 & 74.0 & 56.3\\
DFN-2B & DCI-extended & 77 & 65.7 & 73.5 & 56.1\\
DFN-2B & DCI-complete & 77 & 65.8 & 73.8 & 56.2\\
DFN-2B & Recap-COCO-30K & 77 & 65.1 & 73.5 & 55.5\\
\midrule
DFN-2B & GBC-1M-long & 255 & 64.7 & 72.4 & 55.1\\
DFN-2B & GBC-10M-short-relation & 255 & 65.2 & 73.8 & 55.4\\
DFN-2B & GBC-10M-long & 255 & 64.6 & 71.9 & 54.6\\
DFN-2B & DOCCI & 255 & \cellcolor{lightmintbg} 66.1 & 74.0 & \cellcolor{lightmintbg} 57.2\\
DFN-2B & DCI-extended & 255 & 65.7 & 75.1 & 55.9\\
DFN-2B & DCI-complete & 255 & 65.6 & 74.0 & \cellcolor{lightpumpkin} 56.8\\
\midrule
DFN-2B & 5$\times$2 & 77  & \cellcolor{lightmintbg}65.9$_{0.2}$ & 74.7$_{0.4}$ & 56.3$_{0.2}$ \\
DFN-2B & 10$\times$1 & 77 & \cellcolor{lightmintbg}66.0$_{0.1}$ & 75.1$_{0.6}$ & \cellcolor{lightpumpkin}56.5$_{0.3}$\\
        \bottomrule[1.5pt]
    \end{tabular}}

\end{table}

\textbf{Fine-tuning on MSCOCO38k and 
MSCOCO128k}. We observe that restricting fine-tuning to MSCOCO samples with permissive licenses does not have a negative impact on performance.

\textbf{Ablation on number of synthetic captions and beam search}.
\citep{vasu2024mobileclip} observed that even though one can generate multiple synthetic captions from a CoCa model, their effectiveness saturates at 2 per sample for classification tasks. We observe similar results using a single CoCa model with various sampling strategies.
We explore varying the  generation method and hyperparameters.
Specifically, we used top-p, top-k, and 
beam-search and observed that beam-search results in qualitatively more diverse 
captions, however, we did not observe any improvement in downstream performance 
when utilized for reinforced training.

\textbf{Fine-tuning on GBC1M, GBC12M, DOCCI, DCI, ReCap-COCO30k}. We observe that most fine-tuning datasets underperform MSCOCO fine-tuning or perform on-par within one standard deviation. An exception is fine-tuning on DOCCI results in 0.8\% improvement in average of 38 evaluations which is more than one standard deviation from the MSCOCO-38k results.

\textbf{Effect of context length.} The context length for training CLIP and CoCa models is typically set to 77. We explore training CoCa models to generate longer captions by setting the context length for training and generation to 255. Most results stay within one standard deviation. Recent works have improved the support for long captions in CLIP models with improved loss functions and training strategies~\citep{zhang2024long,zheng2024dreamlip,najdenkoska2024tulip}. We leave extending these modifications to CoCa models for future work.

\textbf{Effect of synthetic caption diversity}. 
We further explore training with a diverse collection of captions generated from an ensemble of CoCa models fine-tuned on different datasets. The motivation is the diversity in fine-tuning datasets would increase the divresity in synthetic captions and hence an increase in the effectiveness of additional synthetic captions. We observe that utilizing up to 10 different CoCa models results in a performance that is still within one standard deviation of the best performance.

\textbf{Reinforced DFN datasets.} Our final datasets small \dfndrFiveTM{} and \dfndrTwoTM{} consist of 5 synthetic captions with MSCOCO-38k fine-tuning, and embeddings from the ensemble of two DFN2B-ViT-L/14 teachers discussed in \cref{sec:method_clip_teachers} for 30 image augmentations as well as ground-truth and synthetic captions. We explored training on only the 2B subset of DFN versus the full 5B set that was expanded with 3B samples outside of the 12B pool of DataComp. \cref{tab:dfn2b_vs_5b} shows that the average performance on 38 evaluations is within standard deviation for both datasets while ImageNet-1k validation accuracy is better with the 12M samples from 5B. However, we did not observe the improvement to hold when training at larger scales and restricted our recipe to the 2B dataset.

\begin{table}[h]
    \centering
    \caption{\textbf{\dfndrFiveTM{} and \dfndrTwoTM{} perform similarly on average 38 evaluations.}}
    \label{tab:dfn2b_vs_5b}
\resizebox{0.4\linewidth}{!}{%
        \resizebox{0.6\columnwidth}{!}{
    \begin{tabular}{c|ccc}
        \toprule[1.5pt]
        Dataset & \imagenetval{}& \flickrval{}& \avgTE{}\\
        \midrule[1.25pt]
DFNDR-5B12M & \cellcolor{lightmintbg} 65.9$_{0.3}$ & \cellcolor{lightmintbg} 75.4$_{0.2}$ & \cellcolor{lightmintbg}56.5$_{0.3}$ \\
DFNDR-2B12M & 65.5 & 74.8 & \cellcolor{lightmintbg}56.4 \\
        \bottomrule[1.5pt]
    \end{tabular}}

}
\end{table}

\section{Architecture}

\begin{figure}
\begin{subfigure}[t]{0.57\linewidth}
    \centering
    \includegraphics[width=\linewidth]{figures/arch_image_encoder.pdf}
    \vspace*{-10pt}
    \caption{\textbf{\mci{} architectures for new larger variants}. For smaller 
    variants, the image encoder has four distinct stages of compute and for 
    larger variants, we use five stages. The projection layer for \mci{} models include a Global Avg. Pooling layer followed by a linear layer.}\label{fig:arch_image_encoder}
\end{subfigure}
\hfill
\begin{subfigure}[t]{0.4\linewidth}
    \centering
    \hspace{-10pt}
    \includegraphics[width=\linewidth]{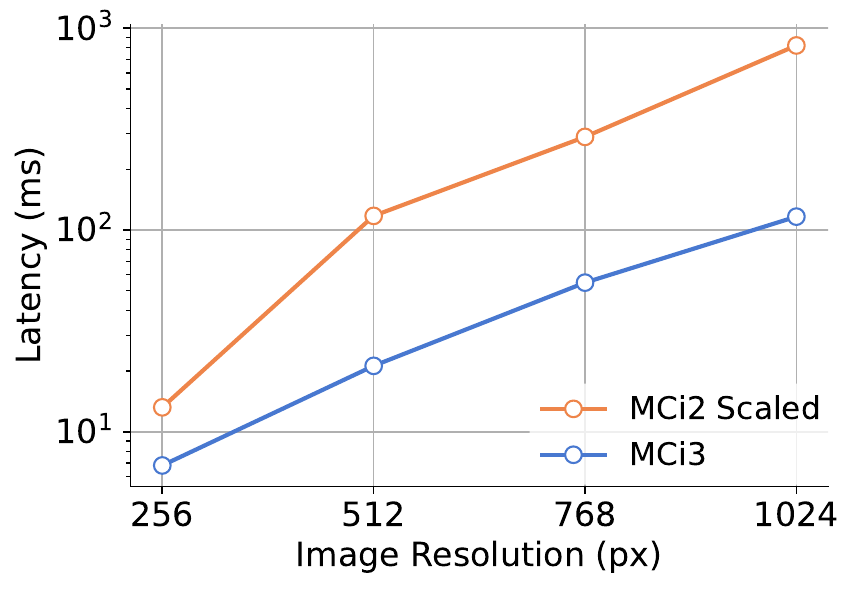}
    \vspace*{-2pt}
    \caption{\textbf{5-Stage achieves lower latency at higher resolutions.} Both models \mciTwo{}-Scaled and \mciThree{} are of the same size, with \mciThree{} using a 5-stage design.}\label{fig:mci3_design}
\end{subfigure}
\caption{\textbf{\ourmodel{} architecture and latency.}}
\end{figure}

Our \ourmodel{} consists of similar architectures to \mobileclip{} as well as two new variants. Specifically, we train \ourmodelSZero{}, \ourmodelSTwo{}, and \ourmodelB{} where we utilize the standard ``Base'' text encoder for \ourmodelSZero{} and drop the S1 variant.
In addition to architectures introduced in \mobileclip{}, we introduce two new 
variants in \ourmodel{} family, i.e., \ourmodelL{} and \ourmodelXL{}. The text 
encoders for these variants are pure transformer-based architectures and the 
image encoders are based on FastViT~\citep{vasu2023fastvit}, which uses train-time overparameterization blocks introduced in~\citep{MobileOne}. The smaller 
variants, \mciZero{}, \mciOne{}, and \mciTwo{} are hybrid vision transformers 
with four distinct stages of compute. We introduce an additional transformer 
stage for \mciThree{} and \mciFour{} preceded by 4$\times$ down-sampling of 
input tensor as shown in \cref{fig:arch_image_encoder}. The 5-stage design has 
two advantages when scaled up; First, the parameters can be distributed across 
five stages with the largest layers operating on four times fewer tokens.  
Second, the design scales more effectively to higher resolutions.

We empirically validate our design choices across various image resolutions. In \cref{fig:mci3_design}, we scale \mciTwo{} to match the size of \mciThree{} (125 million parameters) and benchmark its performance across four input resolutions. Our results show that \mciThree{}, with its five-stage design, offers a significantly better trade-off compared to a scaled \mciTwo{}. At low image resolution, i.e., 256$\times$256, \mciThree{} is 1.9$\times$ faster than similar sized \mciTwo{} and for larger input resolutions, i.e., as 1024$\times$1024,
\mciThree{} is 7.1$\times$ faster than a similar sized \mciTwo{}. Responsiveness at higher resolutions is particularly important when the image encoder is fine-tuned for dense prediction tasks such as image segmentation, where the input image resolution is 512$\times$512.

\section{Experiments}

In this section, we train a new family of efficient CLIP models, \ourmodel{},
and evaluate on a diverse set of tasks.  Following our findings in 
\cref{sec:method}, we create the reinforced dataset, \dfndrTwoB{}, which 
contains five synthetic captions generated from our CoCa-ViT-L/14 model 
pretrained on \dfnTwoB{} and fine-tuned on MSCOCO-38K.  \dfndrTwoB{} also 
contains image-text embeddings from an ensemble of CLIP models, 
\texttt{DFN2B-CLIP-ViT-L-14-s39b} and \texttt{DFN2B-CLIP-ViT-L-14}, for all 
images, ground-truth captions, and synthetic captions. We train a more diverse 
family of architectures compared with MobileCLIP and evaluate their performance 
on 38 zero-shot classification tasks~\citep{gadre2024datacomp}.  Particularly, 
we introduce \ourmodelL{} and \ourmodelXL{} architectures trained on 
\dfndrTwoB{} as well as variants trained on \datacompdrOneB{} which we refer to 
as \mobileclipL{} and \mobileclipXL{}.  \Cref{tab:full_eval} shows our results 
compared with other models with similar latencies. Details of training and 
hyperparameters are described in  \cref{app:hparams}.

We compare \ourmodel{} to prior small CLIP architectures 
TinyCLIP~\citep{tinyclip} trained on LAION~\citep{Laion,Laion400} and 
ACED~\citep{udandarao2024active}.
We also compare with larger models from OpenAI's CLIP~\citep{CLIP},
DataComp~\citep{gadre2024datacomp}, VeCLIP~\citep{VeCLIP}, EVA~\citep{EVA-CLIP},
DFN~\citep{fang2023data}, SigLIP~\citep{SigLIP}, and 
SigLIP2~\citep{tschannen2025siglip}.
We evaluate all models using OpenCLIP~\citep{OpenCLIP} and 
DataComp~\citep{gadre2024datacomp}. In some cases such as SigLIP2, we observe 
positive/negative gaps with reported results in their paper.

\ourmodel{} achieves state-of-the-art ImageNet-1k validation zero-shot 
accuracies at various latencies. Notably, \ourmodelXL{} matches the zero-shot 
accuracy of SigLIP-SO400M/14 on ImageNet validation set while being 2$\times$ 
smaller and improves on DFN ViT-L/14 at 2.5$\times$ lower latency.
We also improve on ImageNet-1k performance of ACED models considering their 
latencies. As ACED optimized their models for low inference flops, the latency 
of both ACED-F1 and ACED-F2 are comparable to our \ourmodelSTwo{} architecture 
while still have higher latency and more parameters.
SigLIP-B/16 and SigLIP2-B/16 models are more comparable in size and latency to 
our new larger architectures. Particularly, SigLIP2 models have substantially 
larger text-encoders compared to SigLIP models.

We note that our models pretrained on \dfndrTwoB{} do not always achieve 
state-of-the-art retrieval performance. We attribute this to the bias of 
\dfndrTwoB{} dataset towards zero-shot classification tasks and particularly 
ImageNet-1k. We observe that models trained on DataComp, WebLI, and their 
derivatives may achieve higher retrieval performance compared to DFN datasets 
and derivatives while lower on \avgTE{} performance. As such, we also 
train our new architectures on \datacompdrOneB{} referred to as \mobileclipL{} 
and \mobileclipXL{}. The combination of these two families of architectures 
will provide flexibility for broader applications.

\begin{table*}[t!]
    \centering
    \vspace{-5pt}
    \caption{\textbf{\ourmodel{} family of models has the best
    average performance at various latencies.}
    Retrieval performances are reported @1. 
    Last column shows average performance on 38 datasets as in OpenCLIP~\citep{OpenCLIP}. Models are grouped by their total latency in increasing order and by performance within each group. ``Base'' refers to standard CLIP Transformer-based~\citep{transformers} text encoder with 12 layers, and ``Custom'' stands for customized text encoder used in the respective method. Models with substantially  higher latencies and/or larger model sizes are grayed out. %
    }
    \vspace{-5pt}
    \label{tab:full_eval}
    \resizebox{0.99\textwidth}{!}{
        \begin{tabular}{l|cC{1.7cm}C{1.7cm}|C{2cm}C{1.5cm}C{2.2cm}C{2.3cm}|cccccc|C{1.8cm}}
            \toprule[1.5pt]
            \multirow{2}{*}{\textbf{Name}} 
            & \multirow{2}{*}{\textbf{Dataset}}
            & \multirow{2}{1.7cm}{\centering\textbf{Seen Samples}}
            & \multirow{2}{1.8cm}{\centering\textbf{Resolution}}
            & \multirow{2}{2cm}{\centering\textbf{Image Encoder}}
            & \multirow{2}{1.5cm}{\centering\textbf{Text Encoder}}
            & \multirow{2}{2.2cm}{\centering\textbf{Params (M) {(img+txt)}}}
            & \multirow{2}{2.3cm}{\centering\textbf{Latency (ms) {(img+txt)}}}
            & \multicolumn{2}{c}{\textbf{Zero-shot CLS}}
            & \multicolumn{2}{c}{\textbf{\flickrval{} Ret.}}
            & \multicolumn{2}{c}{\textbf{COCO Ret.}}
            & \multirow{2}{1.8cm}{\centering\textbf{Avg. Perf. on 38}}
            \\
            \cmidrule(lr){9-10}
            \cmidrule(lr){11-12}
            \cmidrule(lr){13-14}
             & & & & & &&&
             \imagenetval{} & IN-shift &T$\to$I&I$\to$T&T$\to$I&I$\to$T& \\
             \midrule[1.25pt]

             TinyCLIP-RN19M & LAION-400M &
             15.2B & 224 & ResNet-19M & Custom & 18.6 + 44.8 & 1.9 + 1.9 & 56.3 & 43.6 & 58.0 & 75.4 & 30.9 & 47.8 & 48.3 \\

             TinyCLIP-RN30M & LAION-400M &
             15.2B & 224 & ResNet-30M & Custom & 29.6 + 54.2 & 2.6 + 2.6 & 59.1 & 45.7 & 61.5 & 80.1 & 33.8 & 51.6 & 50.2 \\

             TinyCLIP-40M/32 & LAION-400M &
             15.2B & 224 & ViT-40M/32 & Custom & 39.7 + 44.5 & 3.0 + 1.9 & 59.8 & 46.5 & 59.1 & 76.1 & 33.5 & 48.7 & 51.2 \\

             \mobileclipSZero{} & \datacompdrOneB{} & 13B &
             256 & \mciZero{}&\mct{} & 11.4 + 42.4 & 1.5 + 1.6 &
             67.8 & 55.1 & 67.7 & 85.9 & 40.4 & 58.7 & 58.1\\
             ACED-F0 & \datacompOneB{} &
             13B & 256 & ViT-S/32 & Small & 22.7 + 28.8 & 2.1 + 1.8 &
             68.5 & (-) & \textbf{71.4} & \textbf{87.6} & 41.2 & 60.8 & (-) \\
             \rowcolor{lightmintbg}
             \textbf{\ourmodelSZero{}} & \dfndrTwoB{} &
             13B & 256 & \mciZero{} & Base
             & 11.4 + 63.4 & 1.5 + 3.3 &
             \textbf{71.5} & \textbf{57.6} & 69.2 & 86.6 & \textbf{43.7} 
             &\textbf{62.7} & \textbf{59.7} \\

             \midrule
             
             OpenAI-RN50 & OpenAI-400M &
             13B & 224 & ResNet-50 & Base & 38.3 + 63.4 & 3.3 + 3.3 & 
             59.8 & 45.1 & 57.4 & 80.0 & 28.5 & 48.8 & 48.1 \\

             TinyCLIP-61M/32 & LAION-400M &
             15.2B & 224 & ViT-61M/32 & Custom & 61.4 + 54.0 & 4.3 + 2.6 & 62.4 & 48.7 & 62.6 & 78.7 & 36.5 & 52.8 & 53.0 \\

             TinyCLIP-63M/32&
             \begin{tabular}{c}
                LAION-400M\\
                YFCC-15M
             \end{tabular}
             &  15.8B & 224 & ViT-63M/32 & Custom & (-) & (-) & 
             64.5 & (-) & 66.0 & 84.9 & 38.5 & 56.9 & (-) \\

             \mobileclipSOne{} & \datacompdrOneB{} &
             13B & 256 &\mciOne{}& Base & 21.5 + 63.4 & 2.5 + 3.3 &
             \textbf{72.6} & \textbf{60.7} & \textbf{71.0} & \textbf{89.2} & \textbf{44.0} & \textbf{62.2} & \textbf{61.3} \\

             \midrule

             OpenAI-RN101 & OpenAI-400M &
             13B & 224 & ResNet-101 & Base & 56.3 + 63.4 & 4.3 + 3.3 & 
             62.3 & 48.5 & 58.0 & 79.0 & 30.7 & 49.8 & 50.3 \\

             OpenAI-B/32 & OpenAI-400M &
             13B & 224 & \multirow{3}{*}{ViT-B/32} & \multirow{3}{*}{Base} 
             & \multirow{3}{*}{86.2 + 63.4} & \multirow{3}{*}{5.9 + 3.3} & 
             63.3 & 48.5&58.8&78.9&30.4&50.1&52.5\\
             LAION-B/32 & LAION-2B &
             32B & 224 & & & & & 
             65.7 & 51.9&66.4&84.4&39.1&56.2&54.8\\
             DataComp-B/32 & \datacompOneB{} &
             13B & 224 & & & & & 
             69.2 & 55.2&61.1&79.0&37.1&53.5&58.0\\
             
             DataComp-B/32-256 & \datacompOneB{} 
             &  34B & 256 & ViT-B/32 & Base & 86.2 + 63.4 & 6.2 + 3.3 & 
             72.8 & 58.7&64.9&84.8&39.9&57.9&60.9\\

             SigLIP2-B/32 & WebLI-10B
             & 40B & 256 & ViT-B/32 & Custom & 94.6 + 282.3 & 6.3 + 6.3 &
             73.8 & 57.8 & 73.2 & 88.0 & 47.9 & 64.9 & 61.9 \\

             \mobileclipSTwo{} &  \datacompdrOneB{}  &
             13B & 256 & \mciTwo{} & Base & 35.7 + 63.4 & 3.6 + 3.3 &
             74.4 & 63.1 & 73.4 & 90.3 & 45.4 & 63.4 & 63.7 \\
             ACED-F1 & \datacompOneB{} &
             13B & 256 & ViT-B/32 & Small & 86.2 + 28.8  & 6.2 + 1.8  &
             74.9 & (-) & 77.9 & 90.3 & 47.3 & 74.9 & (-) \\
             ACED-F2 & \datacompOneB{} &
             13B & 256 & ViT-B/24 & Small & 86.2 + 28.8 & 6.5 + 1.8 & 
             76.9 & (-) & \textbf{79.5} & \textbf{91.1} & \textbf{49.7} & \textbf{66.9} & (-) \\
             \rowcolor{lightmintbg}
             \textbf{\ourmodelSTwo{}} &  \dfndrTwoB{}  &
             13B & 256 & \mciTwo{} & Base & 35.7 + 63.4 & 3.6 + 3.3 &
             \textbf{77.2} & \textbf{64.7} & 74.8 & 90.4 & 48.8 & \underline{66.7} & \textbf{64.1} \\

             \midrule
             VeCLIP-B/16 &  WIT-200M&
             6.4B & 224 &  \multirow{9}{*}{ViT-B/16} & Base & 86.2 + 63.4 & 11.5 + 3.3 & 
             64.6 & (-) & 76.3 & 91.1 & 48.4 & 67.2 & (-) \\
             OpenAI-B/16 &  WIT-400M &
             13B & 224 & & Base & 86.2 + 63.4 & 11.5 + 3.3 & 68.3 & 55.9 & 67.7 & 85.9 & 40.4 & 58.7 & 58.1 \\
             LAION-B/16 &  LAION-2B &
             34B & 224 &  & Base & 86.2 + 63.4 & 11.5 + 3.3 & 
             70.2 & 56.6 & 69.8 & 86.3 & 42.3 & 59.4 & 58.7\\
             EVA02-B/16 & Merged-2B &
             8B & 224 & & Base & 86.2 + 63.4 & (-) & 
             74.7 & 59.6 & 71.5 & 86.0 & 42.2 & 58.7 & 58.9\\
             DFN-B/16 &  DFN-2B &
             13B & 224 &   & Base & 86.2 + 63.4 & 11.5 + 3.3 & 
             76.2 & 62.3 & 69.1 & 85.4& 43.4 & 60.4& 60.9\\
             DataComp-B/16 & \datacompOneB{} &
             13B & 224 &  & Base & 86.2 + 63.4 & 11.5 + 3.3 & 
             73.5 & 60.8 & 69.8 & 86.3& 42.3 & 59.4& 61.5\\
             
             \mobileclipB{} & \datacompdrOneB{} &
             13B & 224 &   & Base & 86.3 + 63.4 & 10.4 + 3.3 &
             76.8 & 65.6 & \textbf{77.3} & 91.4& \textbf{50.6} & \textbf{68.8} & 65.2\\
             \mobileclipB{} (LT) & \datacompdrOneB{} &
             39B & 224 &   & Base & 86.3 + 63.4 & 10.4 + 3.3 &
             77.2 & 66.1 & 76.9 & \textbf{92.3} & 50.0 & \underline{68.7} & \textbf{65.8}\\
             \rowcolor{lightmintbg}
             \textbf{\ourmodelB{}} & \dfndrTwoB{} &
             13B & 224 &   & Base & 86.3 + 63.4 & 10.4 + 3.3 &
             \textbf{79.4} & \textbf{66.4} & 76.5 & 89.7 & 49.9 & 67.5 
             & \textbf{65.8} \\

             \midrule
             SigLIP-B/16 & WebLI &
             40B & 224 & ViT-B/16 & Custom & 92.9 + 110.3 & 9.9 + 5.8 &
             76.0 & 61.0 & 74.7 & 89.1& 47.8 & 65.7& 62.3\\

             SigLIP-B/16-256 & WebLI &
             40B & 256 & ViT-B/16 & Custom & 92.9 + 110.3 & 11.4 + 5.8 &
             76.5 & 62.0 & 75.0 & 90.4& 48.4 & 66.1& 62.3\\
             SigLIP2-B/16 & WebLI-10B &
             40B & 224 & ViT-B/16  & Custom & 92.9 + 282.3 & 9.9 + 6.3 &
             78.5 & 63.9 & 79.3 & 93.1 & 53.2 & 69.4 & 64.6\\

             SigLIP2-B/16-256 & WebLI-10B &
             40B & 256 & ViT-B/16  & Custom & 92.9 + 282.3 & 11.4 + 6.3 &
             79.3 & 65.3 & \textbf{80.2} & 93.2 & \textbf{54.1} & \textbf{70.8} 
                & 64.6\\
             \rowcolor{lightmintbg}
             \textbf{\mobileclipL{}} & \datacompdrOneB{} &
             13B & 256 &  \mciThree{} & Large &
             125.1 + 123.6 & 8.0 + 6.6 &
             78.3 &  68.2 & \textbf{77.9} & \textbf{93.1} & \textbf{51.3} & \textbf{68.8} & 66.3 \\
             \rowcolor{lightmintbg}
             \textbf{\ourmodelL{}} & \dfndrTwoB{} &
             13B & 256 &  \mciThree{} & Large &
             125.1 + 123.6 & 8.0 + 6.6 &
             \textbf{80.7} &  \textbf{68.9} & 77.3 & 91.6 & 50.9 & 68.4 & \textbf{66.8} \\
             \midrule
             SigLIP-L/16 & WebLI &
             40B & 256 & ViT-L/16 & Custom & 316.0 + 336.2 & 38.2 + 19.1 &
             80.4 & 66.6 & 79.0 & 91.8 & 52.3 & \textbf{70.8} & 65.6 \\
             DFN-L/14-quickgelu & \dfnTwoB{} &
             13B & 224 & ViT-L/14 & Large & 304.3 + 123.6 & 57.9 + 6.6 &
             81.4 & 68.8 & 78.5 & 89.0 & \textbf{53.7} & 66.8 & 66.9 \\
             \rowcolor{lightmintbg}
             \textbf{\mobileclipLFourteen{}} & \datacompdrOneB{} &
             13B & 224 & ViT-L/14  & Large & 304.3 + 123.6 & 57.9 + 6.6 &
             79.5 & 69.9 & 75.3 & 91.3 & 47.6 & 66.5 & 66.9 \\
             \rowcolor{lightmintbg}
             \textbf{\ourmodelXL{}} & \dfndrTwoB{} &
             13B & 256 & \mciFour{}  & Large &
             321.6 + 123.6 & 19.6 + 6.6 &
             \textbf{81.9} &  \textbf{70.3} & 78.0 & 92.4 & 51.5 & 69.3 &  67.5 \\
             \rowcolor{lightmintbg}
             \textbf{\ourmodelLFourteen{}} & \dfndrTwoB{} &
             13B & 224 & ViT-L/14  & Large & 304.3 + 123.6 & 57.9 + 6.6 &
             \textbf{81.9} & \textbf{70.2} & 77.2 & 92.0 & 51.6 & 69.0 & \textbf{67.8} \\
             \rowcolor{lightmintbg}
             \textbf{\mobileclipXL{}} & \datacompdrOneB{} &
             13B & 256 & \mciFour{}  & Large &
             321.6 + 123.6 & 19.6 + 6.6 &
             79.4 &  69.7 & \textbf{79.5} & \textbf{94.9} & 52.1 & 70.3 & \textbf{68.1} \\
             \midrule
             \rowcolor{lightpumpkin}
             SigLIP-SO400M/14 & WebLI &
             40B & 224 & So-400M & Custom &
             427.7 + 449.7 & (-) &
             82.0 & 69.5 & 75.2 & 91.0 & 51.8 & 69.7 & 68.1 \\
             \rowcolor{lightpumpkin}
             SigLIP2-L/16 & WebLI-10B &
             40B & 256 & ViT-L/16 & Custom & 316.0 + 565.6 & 38.2 + 19.8 &
             82.3 & 70.5 & 81.8 & 94.6 & 54.7 & 72.0 & 68.3 \\
             \rowcolor{lightpumpkin}
             SigLIP2-SO400M/14 & WebLI-10B &
             40B & 224 & So-400M & Custom &
             427.7 + 707.8 & (-) &
             83.2 & 72.0 & 82.8 & 93.9 & 55.5 & 71.9 & 69.1 \\
             
            \bottomrule[1.5pt]
        \end{tabular}
    }
    \vspace{-5pt}
\end{table*}

\subsection{VLM evaluations}

We report vision-language evaluations using \ourmodel{} pretrained models in 
the LLaVA-{1.5} setup~\citep{liu2024improved}. We keep the vision backbone 
frozen for all the runs and use Qwen2-7B instead of Vicuna-7B. All other 
training details are the same as the original LLaVA-1.5 setup, more details are provided in appendix. We evaluate 
ViT-B/16 models pretrained on DataComp, DFN, DataCompDR, and DFNDR for 13B seen 
samples. In \cref{tab:llava_evals} we observe that on average training on DFNDR achieves {3.5\%} higher 
accuracy compared with DFN pretrained model, {1.6\%} better than DataComp 
pretrained model, and {0.6\%} better than DataCompDR pretrained model. 

\begin{table}[t]
    \centering
            \caption{\textbf{VLM evaluations in LLaVA-1.5 setup.} ViT-B/16 pretrained models reach \textbf{{3.5\%} higher accuracy} compared with DFN pretrained model, {1.6\%} better than DataComp 
pretrained model, and {0.6\%} better than DataCompDR pretrained model.}\label{tab:llava_evals}
        \resizebox{0.7\linewidth}{!}{
            \begin{tabular}{l|cccccccc|c}
                \toprule[1.5pt]
                \textbf{Dataset} & \textbf{GQA} & \textbf{SQA} 
                & \textbf{TextVQA} & \textbf{POPE} & \textbf{MMMU} 
                & \textbf{MMB} & \textbf{VizWiz} & \textbf{VQAv2} 
                & \textbf{Avg.} \\
\midrule
\datacompOneB{}   & 59.6          & 71.5          & \textbf{50.5} & 81.8          & 42.6          & 59.1          & 51.8          & 70.7          & 61.0          \\
\dfnTwoB{}        & 56.9          & 71.3          & 46.0          & 81.4          & 41.9          & 52.2          & \textbf{56.1} & 66.9          & 59.1          \\
\datacompdrOneB{} & 60.3          & \textbf{73.1} & 50.4          & 81.7          & 43.6          & 60.2          & 54.9          & 72.1          & 62.0          \\
\dfndrTwoB{}      & \textbf{60.4} & 72.9          & 49.9          & \textbf{83.3} & \textbf{45.2} & \textbf{61.9} & 54.5          & \textbf{72.4} & \textbf{62.6} \\
                \bottomrule[1.5pt]
        \end{tabular}
        }
\end{table}

\subsection{Dense Prediction tasks}
We evaluate the quality of the visual representations learned by finetuning the image encoder on dense prediction tasks like object detection, semantic segmentation and depth estimation. In~\Cref{tab:maskrcnn_results}, we report performance of ViT-B/16 model with MaskRCNN~\cite{maskrcnn} head for instance segmentation on MS-COCO~\cite{MSCOCO} dataset. All models were trained using MMDetection library~\cite{mmdetection} using 1$\times$ schedule with single scale testing as described in~\cite{clipft}. We follow finetuning setup described in~\cite{clipft}, more details in appendix. In Table~\Cref{tab:segm_results}, we report performance of ViT-B/16 model with UperNet~\cite{upernet} head, trained using the same setup described in~\cite{dBOT} on ADE20k~\citep{ADE20K} dataset. In~\Cref{tab:depth_results}, we report Root Mean Square Error (RMSE) on NYUv2 dataset~\cite{nyuv2}.
We use the same settings as described in ~\cite{vasu2024qualitycaptions}, more details are provided in appendix.

Additionally, we assess the performance of smaller \ourmodel{} variants on dense prediction tasks. Popular pretraining methods like MAE~\citep{mae}, are not directly applicable to hierarchical convolutional and hybrid architectures such as our MCi models, hence we compare \ourmodel{} pretraining with supervised pretraining for the same architectures. In \cref{tab:mobile_det_results,tab:mobile_segm_results}, we see that \ourmodel{} pretraining is significantly better than supervised pretraining and can serve as a good pretraining choice for hierarchical architectures.

\begin{minipage}[t!]{0.47\textwidth}%
        \centering
        \captionof{table}{Object detection and instance segmentation results on MS-COCO with Mask-RCNN head trained for 1$\times$ schedule. All models are ViT-B/16.}\label{tab:maskrcnn_results}
        \resizebox{0.98\linewidth}{!}{
            \begin{tabular}{l@{\hspace*{5mm}} c@{\hspace*{5mm}}c@{\hspace*{5mm}}c}
                \toprule[1.5pt]
                \textbf{Method} & \textbf{Dataset} & \textbf{mAP}$^{box}$ & \textbf{mAP}$^{mask}$ \\
                \midrule
                CatLIP~\cite{mehta2024catlip} & \datacomp{} & 45.7 & 40.6 \\
                MAE~\cite{mae}  & IN-1K & 46.5 & 40.9 \\
                MAE~\cite{singh2023effectiveness}  & IG-3B & 46.4 & 42.1 \\
                MAWS~\cite{singh2023effectiveness} & IG-3B & 48.0 & \textbf{43.4} \\
                FD-CLIP~\cite{clipft}         & OpenAI-WIT + IN-1K & \textbf{48.2} & 42.5 \\
                \midrule[0.5pt]
                CLIP~\cite{CLIP}            & OpenAI-WIT & 45.0 & 39.8 \\
                \textbf{\ourmodel{}}  & \dfndrTwoB{} & 47.0 & 41.8 \\
               \bottomrule[1.5pt]
        \end{tabular}
        }
\end{minipage}%
\hspace{10px}
\begin{minipage}[t!]{0.47\textwidth}%
        \centering
        \captionof{table}{Semantic segmentation results on ADE20k using UperNet decoder. All models are ViT-B/16.}\label{tab:segm_results}
        \resizebox{0.9\linewidth}{!}{
            \begin{tabular}{l@{\hspace*{5mm}} c@{\hspace*{5mm}}c@{\hspace*{5mm}}c}
                \toprule[1.5pt]
                \textbf{Method} & \textbf{Dataset} & \textbf{mIoU} & \textbf{mAcc} \\
                \midrule
                MAE~\cite{mae}                     & IN-1K & 48.1 & 58.9 \\
                dBOT~\cite{dBOT} & IN-1K        & 49.5 & 60.7 \\
                MAWS~\cite{singh2023effectiveness} & IG-3B & 50.4 & 61.5 \\
                CatLIP~\cite{mehta2024catlip}   & \datacomp{} & 50.6 & 61.8 \\
                FD-CLIP~\cite{clipft}  & OpenAI-WIT + IN-1K & 51.7 & - \\
                \midrule[0.5pt]
                CLIP~\cite{CLIP}     & OpenAI-WIT         & 49.5 & - \\ 
                \textbf{\ourmodel{}} & \dfndrTwoB{} & \textbf{52.8} & \textbf{64.0} \\
               \bottomrule[1.5pt]
        \end{tabular}
        }
\end{minipage}%

\begin{table}[t!]%
        \centering
        \captionof{table}{Results on NYUv2 for depth estimation following the same settings as~\cite{clipft}. All results are for ViT-B/16 models.}\label{tab:depth_results}
        \resizebox{0.5\linewidth}{!}{
            \begin{tabular}{l@{\hspace*{5mm}} c@{\hspace*{5mm}}c}
                \toprule[1.5pt]
                \textbf{Method} & \textbf{Dataset} & \textbf{RMSE}($\downarrow$) \\
                \midrule
                CatLIP~\cite{mehta2024catlip} & \datacomp{} & 0.394 \\
                MAE~\cite{mae}     & IN-1K & 0.383 \\
                MAWS~\cite{singh2023effectiveness} & IG-3B & 0.371 \\
                FD-CLIP~\cite{clipft} & OpenAI-WIT + IN-1K & 0.352 \\
                MAE~\cite{singh2023effectiveness}  & IG-3B & \textbf{0.348} \\
                \midrule[0.5pt]
                CLIP~\cite{CLIP}    & OpenAI-WIT  & 0.416 \\
                \textbf{\ourmodel{}}    & \dfndrTwoB{} & 0.356 \\
               \bottomrule[1.5pt]
        \end{tabular}
        }
\end{table}%

\begin{table}[t]
    \centering
            \caption{Comparison pretraining methods for semantic segmentation on ADE-20k. For reference, we have included recent state-of-the-art semantic segmentation models (in gray). 
        }\label{tab:mobile_segm_results}
        \resizebox{0.8\linewidth}{!}{
            \begin{tabular}{l@{\hspace*{5mm}} c@{\hspace*{5mm}}c@{\hspace*{5mm}}c@{\hspace*{5mm}}c@{\hspace*{5mm}}c}
                \toprule[1.5pt]
                \textbf{Encoder} & \textbf{Decoder} & \textbf{Pre-Training} & \textbf{Resolution} & \textbf{\# Params}(M) & \textbf{mIoU} \\
                \midrule

                \textcolor{gray}{InternImage-B}~\cite{internimage} & \textcolor{gray}{UperNet}~\cite{upernet} & \textcolor{gray}{Sup. IN-1K} &  \textcolor{gray}{512$\times$512} & \textcolor{gray}{128.0} & \textcolor{gray}{50.8} \\ 
                \textcolor{gray}{ViT-Adapter-B}~\cite{vitadapter} & \textcolor{gray}{SemanticFPN}~\cite{kirillov2019panoptic} & \textcolor{gray}{Sup. IN-22K} &  \textcolor{gray}{512$\times$512} & \textcolor{gray}{104.6} & \textcolor{gray}{50.7} \\
                \textcolor{gray}{ViT-Adapter-B}~\cite{vitadapter} & \textcolor{gray}{UperNet}~\cite{upernet} & \textcolor{gray}{Sup. IN-22K} &  \textcolor{gray}{512$\times$512} & \textcolor{gray}{133.9} & \textcolor{gray}{51.9} \\
                \textcolor{gray}{Swin-L}~\cite{swin} & \textcolor{gray}{UperNet}~\cite{upernet} & \textcolor{gray}{Sup. IN-22K} &  \textcolor{gray}{640$\times$640} & \textcolor{gray}{234.1} & \textcolor{gray}{52.1} \\

                \midrule[0.5pt]

                MCi0 & SemanticFPN~\cite{kirillov2019panoptic} & Sup. IN-1K & 512$\times$512 & 14.5 & 44.8 \\
               
                MCi2 & SemanticFPN~\cite{kirillov2019panoptic} & Sup. IN-1K & 512$\times$512 & 38.5 & 48.9 \\

                \midrule[0.5pt]
                
                MCi0 & SemanticFPN~\cite{kirillov2019panoptic} & \textbf{\ourmodel{}} & 512$\times$512 & 14.5 & 47.0 \textcolor{darkgreen}{(+2.2)} \\

                MCi2 & SemanticFPN~\cite{kirillov2019panoptic} & \textbf{\ourmodel{}} & 512$\times$512 & 38.5 & 51.6 \textcolor{darkgreen}{(+2.7)} \\
                \bottomrule[1.5pt]
        \end{tabular}
        }
\end{table}

\begin{table}[t]
    \centering
            \caption{Comparison pretraining methods for object detection task on MS-COCO using MaskRCNN~\cite{maskrcnn} detection head. All models are trained for 1$\times$ schedule. For reference we have included recent state-of-the-art object detection models (in gray). 
        }\label{tab:mobile_det_results}
        \resizebox{0.7\linewidth}{!}{
            \begin{tabular}{l@{\hspace*{5mm}}c@{\hspace*{5mm}}c@{\hspace*{5mm}}c@{\hspace*{5mm}}c}
                \toprule[1.5pt]
                \textbf{Model} & \textbf{Pre-Training} & \textbf{\# Params}(M) & \textbf{mAP}$^{box}$ & \textbf{mAP}$^{mask}$ \\
                \midrule
                
                \textcolor{gray}{ViT-Adapter-B}~\cite{vitadapter} & \textcolor{gray}{Sup. IN-1K} & \textcolor{gray}{284} & \textcolor{gray}{47.0} & \textcolor{gray}{41.8} \\
                \textcolor{gray}{InternImage-B}~\cite{internimage} & \textcolor{gray}{Sup. IN-1K} & \textcolor{gray}{115} & \textcolor{gray}{48.8} & \textcolor{gray}{44.0} \\
                \textcolor{gray}{ViT-Adapter-L}~\cite{vitadapter} & \textcolor{gray}{Sup. IN-22K} & \textcolor{gray}{347.9} & \textcolor{gray}{48.7} & \textcolor{gray}{43.3} \\
                
                \midrule[0.5pt]
                MCi0 & Sup. IN-1K & 31.0 & 41.8 & 38.0 \\

                MCi2 & Sup. IN-1K & 55.0 & 46.6 & 41.7 \\
              
                \midrule[0.5pt]
                MCi0 & \textbf{\ourmodel{}} & 31.0 & 44.4 \textcolor{darkgreen}{(+2.6)} & 39.6 \textcolor{darkgreen}{(+1.6)} \\
                
                MCi2 & \textbf{\ourmodel{}} & 55.0 & 49.1 \textcolor{darkgreen}{(+2.5)} & 43.2 \textcolor{darkgreen}{(+1.5)} \\
             
                \bottomrule[1.5pt]
        \end{tabular}
        }
\end{table}

\section{Related Work}

Improving the training of multi-modal models focus on three aspects: data, 
objective function and architecture. Our \ourmodel{} builds on MobileCLIP and 
provides improvements in all three aspects.

Data approaches either filter a dataset or augment it with additional 
information. Basic filtering methods begin by selecting or crawling a large 
dataset of candidate image-text pairs and filter using ad-hoc rules based on 
the URLs or statics of the images and 
captions~\citep{CLIP,Laion400,Laion,xu2023demystifying}.  More advanced 
filtering methods involve filtering models trained on high-quality data 
utilized to remove low-quality image-text pairs. These methods may utilize 
a pretrained CLIP model~\citep{gadre2024datacomp} or more specialized filtering 
models~\citep{fang2023data}. The challenge with data methods is that the biases 
introduced by ad-hoc rules or pretrained models. For example, most publicly 
available datasets such as DataComp are filtered for English-only data which 
limits the capabilities of models on non-English 
tasks~\citep{carlsson2022cross,nguyen2024multilingual,pouget2024no}.  
Alternatively, pretrained models may be used for active data selection based on 
the sample difficulty~\citep{evans2024data,evans2024bad}. It has also been 
observed that repeating high-quality data achieves higher 
utilization~\citep{goyal2024scaling}.

More broadly, the output of pretrained models can be stored as part of a new 
augmented dataset. For example, various works utilize image-captioning models 
to generate synthetic captions for images in 
a dataset~\citep{ALIP,SynCapt,VeCLIP,liu2024clips,li2024if}. Large language 
models can also be used to rewrite ground-truth 
captions~\citep{fan2023improving} as well as together with text-to-image models 
to generate fully synthetic datasets~\citep{hammoud2024synthclip}. MobileCLIP 
introduced the multi-modal dataset reinforcement where they utilized an 
image-caption model to generate synthetic captions as well as an ensemble of 
large CLIP models to store CLIP embeddings for multiple image augmentations and 
synthetic captions and store them efficiently~\citep{vasu2024mobileclip}. We 
follow a similar approach while improving both the caption generator and CLIP 
embedding generators through better DFN models~\citep{fang2023data}.

Another approach is to improve 
the objective function of multi-modal training. The original CLIP paper utilized a contrastive loss 
that encourages the representations of images and texts paired in the dataset to be kept close to each other while staying farther away from other images and texts in a mini-batch~\citep{CLIP}.  
SigLIP introduced a variant based on Sigmoid instead of Softmax that achieves 
higher training efficiency at larger batch 
sizes~\citep{SigLIP,tschannen2025siglip}. Other methods utilize objectives 
based on image masking~\citep{RILS,EVA,EVA-CLIP,FLIP} and unimodal 
self-supervision~\citep{SLIP,DeCLIP} as well as multi-resolution 
training~\citep{CLIPA} for cost-effective training.  
Multi-modal distillation achieves more significant improvements, particularly 
for smaller architecture variants~\citep{CLIP-TD,DLIP,MAD,tinyclip}.  Notably, 
MobileCLIP~\citep{vasu2024mobileclip} achieved high training efficiency by 
utilizing an offline knowledge distillation 
method~\citep{shen2022fast,yun2021re,faghri2023reinforce}. We utilize a similar objective 
function as MobileCLIP that includes embedding distillation on image-text pairs 
and synthetic captions.

Lastly, architectural improvements seek improved inference 
efficiency and higher performance given a parameter, flops, or latency budget.  
CLIP architectures are often borrowed from uni-modal image and text models.  
Particularly, the original CLIP and various followup works utilized standard 
ViT architectures together with a modified BERT text 
encoder~\citep{ViT,devlin2019bert,CLIP}. Efficient architectures for CLIP 
include TinyCLIP that prunes ViT~\citep{tinyclip}, \citet{pumer} 
that reduce tokens, and \citet{evans2024bad} that reduce the parameters for 
lower flops.  MobileCLIP introduced efficient architectures specifically design 
for CLIP where they introduced a low latency convolution-transformer hybrid 
architectures for both their image and text encoders. We further improve on 
their architectures by introducing two new variants that fill the large latency 
gap between common B and L architectures.

\section{Conclusion}

We introduce \ourmodel{}, a new family of low latency image-text models, 
achieving state-of-the-art ImageNet-1k zero-shot validation accuracy.
Our methodology improves multi-modal reinforced training by utilizing stronger 
CLIP teachers as well as our newly trained image-captioning models. We 
particularly perform a comprehensive study of tuning and ensembling CLIP 
teachers as well as training and fine-tuning efficient image-captioning models.
Notably, \ourmodelXL{} matches the zero-shot accuracy of SigLIP-SO400M/14 on 
ImageNet-1k while being 2$\times$ smaller and improves on DFN ViT-L/14 at 
{2.5$\times$} lower latency. We release our model checkpoints and data 
generation code that facilitates dataset generation at scale.

\subsubsection*{Broader Impact Statement}

Our work introduces a family of foundation models particularly optimized for 
deployment on mobile and edge devices. As such, it facilitates broader use of 
foundation models and development of applications for wider user bases.  
\ourmodel{} may be used for various applications such as image classification 
where its output is impacted by the existing biases of the training datasets 
and teacher models.

\subsubsection*{Acknowledgments}
We would like to thank Albin Madappally Jose, Barry Theobald, Chen Huang, Rick Chang, and Apple Machine Learning 
Research team for their help and discussions throughout this project.

\bibliography{main}
\bibliographystyle{tmlr}

\clearpage
\appendix

\section{Experimental Setup}\label{app:hparams}
\Cref{tab:dataset_summary} provides a summary of datasets used in our ablations and experiments.

\begin{table}[h]
    \centering
    \caption{\textbf{Summary of pretraining datasets.}
    \label{tab:dataset_summary}
    }
        \resizebox{0.99\textwidth}{!}{
    \begin{tabular}{c|cccccc|c}
        \toprule[1.5pt]
Dataset
& Num. Samples
& CLIP Teachers
& Syn. Captioner
& Num. Image Augs.
& Num. Syn. Caps.
& BFloat16
& Size (TBs)
\\
        \midrule[1pt]
\datacompTM{}      & 12.8M  & \xmark                &  \xmark                             & \xmark & \xmark & \xmark & 0.9 \\
\dfnTwoTM{}        & 12.8M  & \xmark                &  \xmark                             & \xmark & \xmark & \xmark & 0.2 \\
\dfnFiveTM{}       & 12.8M  & \xmark                &  \xmark                             & \xmark & \xmark & \xmark & 1.5 \\  %
\datacompdrTM{}    & 12.8M  & OpenAI + DataComp-XL  &  LAION-2B $\rightarrow$ MSCOCO-123k & 30     & 5      & \cmark & 2.1 \\
\dfndrTwoTM{}      & 12.8M  & DFN-2B + DFN-2B-s39B  &  DFN-2B $\rightarrow$ MSCOCO-38k    & 30     & 5      & \cmark & 1.3 \\ %
\dfndrFiveTM{}     & 12.8M  & DFN-2B + DFN-2B-s39B  &  DFN-2B $\rightarrow$ MSCOCO-38k    & 30     & 5      & \cmark & 2.6 \\ %
\midrule[1.25pt]     
\datacompOneB{}    &  1.3B  & \xmark                &  \xmark                             & \xmark & \xmark & \xmark & 90  \\
\dfnTwoB{}         &  1.9B  & \xmark                &  \xmark                             & \xmark & \xmark & \xmark & 65  \\ %
\datacompdrOneB{}  &  1.3B  & OpenAI + DataComp-XL  &  LAION-2B $\rightarrow$ MSCOCO-123k & 10     & 5      & \xmark & 140 \\
\dfndrTwoB{}       &  1.9B  & DFN-2B + DFN-2B-s39B  &  DFN-2B $\rightarrow$ MSCOCO-38k    & 2      & 5      & \xmark & 162 \\ 
        \bottomrule[1.5pt]
    \end{tabular}}
\end{table}

\Cref{tab:hyperparams_v2} summarizes the hyperparameters we used to train 
\ourmodel{}.
For training on 13B seen samples, we use either a setup with 32x8xA100-40GB 
GPUs or a setup with 16x8xH100-80GB GPUs. For our ablations we train for 30k 
seen samples using 4x8xH100-80GB GPUs and global batch size 8192.

\begin{table}[h]
\centering
\caption{Training hyperparameters for our CLIP experiments on \dfndrTwoB{}.}
\scalebox{0.85}{
\begin{tabular}{l|C{1.5cm}|C{1.5cm}|C{1.5cm}|C{1.5cm}|C{1.5cm}}
\toprule
    Hyperparameter   & S0 & S2 & B & S3 & S4 \\
    \midrule
    Input resolution & 256$^2$ &  256$^2$ & 224$^2$ & 256$^2$ & 256$^2$\\
    Context length & \multicolumn{5}{c}{77}\\
    Data augmentation & \multicolumn{5}{c}{RandAugment}  \\
    Random resize crop scale & \multicolumn{5}{c}{[0.08, 1.0]} \\
    Random resized crop ratio & \multicolumn{5}{c}{[0.75, 1.33]} \\
    RangeAugment target value & \multicolumn{5}{c}{(40, 20)} \\
    Train iterations & \multicolumn{5}{c}{200k} \\
    Warmup iterations & 10k & 10k & 2k & 2k & 2k\\
    Global batch size & 65536 & 65536 & 65536& 114688 & 114688\\
    Optimizer & \multicolumn{5}{c}{AdamW} \\
    AdamW beta1 & \multicolumn{5}{c}{0.9}\\
    AdamW beta2 & \multicolumn{5}{c}{0.95}\\
    Max learning rate & \multicolumn{5}{c}{1e-3} \\
    Min learning rate & 1e-6 & 1e-6 & 1e-6 & 0 & 0 \\
    LR. decay schedule & \multicolumn{5}{c}{cosine} \\
    Weight decay rate & \multicolumn{5}{c}{0.2} \\
    Gradient clipping & \multicolumn{5}{c}{1.0} \\
    Mixed precision& \multicolumn{5}{c}{BFloat16} \\
    EMA decay rate & 0.9995 & No EMA & No EMA & No EMA & No EMA \\
    CLIP loss weight & 0.0 & 0.0 & 0.0 & 0.0 & 0.0\\
    KD loss weight & 1.0 & 1.0 & 1.0 & 1.0 & 1.0\\
    GT caption weight & \multicolumn{5}{c}{1.0}\\
    Synth.\ caption weight & \multicolumn{5}{c}{1.0}\\
    Synth.\ teacher & \multicolumn{5}{c}{CoCa-ViT-L/14 - DFN-2B $\rightarrow$ MSCOCO-38k}\\
    Teacher 1 & \multicolumn{5}{c}{DFN2B-CLIP-ViT-L-14-s39b}\\
    Teacher 2 & \multicolumn{5}{c}{DFN2B-CLIP-ViT-L-14} \\
    Teacher 1 logit scale & \multicolumn{5}{c}{70.0} \\
    Teacher 2 logit scale & \multicolumn{5}{c}{60.0} \\
    Teacher resolution & \multicolumn{5}{c}{224$\times$224}\\
\bottomrule

\end{tabular}
}
\label{tab:hyperparams_v2}
\end{table}

\subsection{Training details for CoCa caption generators}
We use OpenCLIP to train CoCa-ViT-L/14 architecture (\texttt{coca\_ViT-L-14}).  
We pretrain models on \dfnTwoB{} and fine-tune on various datasets.  
\Cref{tab:hyperparams_coca} summarizes the hyperparameters for our CoCa 
pretraining and fine-tuning.

\begin{table}[h]
\centering
\caption{Training hyperparameters for our CoCa models trained on \dfnTwoB{}.}
\scalebox{0.85}{
\begin{tabular}{l|c|c}
\toprule
    Hyperparameter   & \dfnTwoB{} Pretrain & Fine-tune \\
    \midrule
    Input resolution & 224$^2$ &  224$^2$ \\
    Context length & 77 & 77, 255\\
    Seen samples & 12.8B & 12M \\
    Train iterations & 195k & 3k, 6k \\
    Early stop iterations & 143k & N/A \\
    Warmup iterations & 10k & 1k \\
    Global batch size & 65536 & 4092, 2048 \\
    Optimizer & \multicolumn{2}{c}{AdamW} \\
    AdamW beta1 & \multicolumn{2}{c}{0.9}\\
    AdamW beta2 & \multicolumn{2}{c}{0.95}\\
    Max learning rate & 1e-3 & 1e-5 \\
    Min learning rate & \multicolumn{2}{c}{0} \\
    LR. decay schedule & \multicolumn{2}{c}{cosine} \\
    Weight decay rate & 0.2 & 0.1 \\
    Gradient clipping & \multicolumn{2}{c}{1.0} \\
    Mixed precision& \multicolumn{2}{c}{amp} \\
    CoCa caption loss weight & \multicolumn{2}{c}{2.0} \\
    CoCa contrastive loss weight & \multicolumn{2}{c}{1.0} \\
    GPU Setup & 32x8xA100-40GBs & 1x8xH100-80GBs\\
\bottomrule

\end{tabular}
}
\label{tab:hyperparams_coca}
\end{table}

\subsection{Training details for VLM}
To assess the quality of the vision encoders, we adopt the LLaVA-1.5~\citep{liu2024improved} training framework. This framework consists of two stages: (1) projector training, and (2) joint fine-tuning of the projector and the language model on 665K instruction-tuning samples. The hyperparameters used in our experiments are summarized in \Cref{tab:hyp-llava}. For the language model, we use Qwen2-7B-Instruct~\cite{Qwen2-VL} as opposed to Vicuna-7B. In both the stages the vision encoder remains frozen. 

\begin{table}[ht]
\centering
\resizebox{0.45\columnwidth}{!}{%
\begin{tabular}{l|cc}
\toprule
           & Stage-1              & Stage-2              \\ 
\midrule
Data            & LLaVA-1.5 558K & LLaVA-1.5 665k \\
\midrule
Learning Rate              & 1e-3           & 2e-5           \\
Global Batch Size      & 256            & 128            \\
Epochs          & 1              &  1 \\
LR. schedule     & cosine decay   & cosine decay   \\
LR. warmup ratio & 0.03           & 0.03           \\
Optimizer       & AdamW          & AdamW      \\ 
\midrule
Trainable & \multirow{2}{*}{Projector} & Projector + \\
modules & & Language Model \\
\bottomrule   
\end{tabular}
}
\caption{LLaVA-1.5 training setup used in ablations for \Cref{tab:llava_evals}.}
\label{tab:hyp-llava}
\end{table}

\subsection{Training details for dense prediction tasks}
\subsubsection{Object detection}
We train object detection models with MaskRCNN detection heads. Along with detection, these models also perform instance segmentation. We follow the settings prescribed in recent works like~\cite{dBOT, clipft, singh2023effectiveness, vasu2024qualitycaptions}. All evaluations reported in the main paper are from single-scale evaluations on MS COCO validation set following prior works. We sweep through stochastic depth rate in steps of \texttt{0.05} and peak learning rate for all the results reported in the paper and the ranges are listed in~\Cref{tab:hyperparams_det_coco}. For ViT-B/16 models, we use ViTDet style feature pyramid network. For MCi architectures, we follow the setup described in ~\cite{vasu2023fastvit}. All models were trained using MMDetection library~\cite{mmdetection} on a single node with 8 A100 NVIDIA GPUs. 

\subsubsection{Semantic Segmentation}
We train segmentation models with UperNet and SemanticFPN heads. These models are trained on ADE20k~\cite{ADE20K} dataset following the settings prescribed in~\cite{dBOT, clipft, singh2023effectiveness, vasu2024qualitycaptions}. All evaluations reported in the main paper are from single-scale evaluations on validation set following prior works. For ViT-B/16 models, we use ViTDet style feature pyramid network with UperNet head. For MCi architectures, we follow the setup described in ~\cite{vasu2023fastvit} and train models with only SemanticFPN head. We sweep through stochastic depth rate in steps of \texttt{0.05} and peak learning rate for all the results reported in the paper and the ranges are listed in~\Cref{tab:hyperparams_segm}. All models were trained using MMSegmentation library~\cite{mmseg2020} on a single node with 8 A100 NVIDIA GPUs.

\subsubsection{Depth Estimation}
We follow the experimental setup and architecture as described in~\cite{clipft, vasu2024qualitycaptions}. The models are trained and evaluated on NYUv2 dataset~\cite{nyuv2}. We sweep through stochastic depth rate in steps of \texttt{0.05} and peak learning rate for all the results reported in the paper and the ranges are listed in~\Cref{tab:hyperparams_depth}.

\begin{minipage}[t!]{0.47\textwidth}%
        \centering
        \captionof{table}{Training hyperparameters for object detection and instance segmentation experiments on MS COCO. ``RRC'' is \texttt{RandomResizedCrop}. We sweep through stochastic depth rate in steps of \texttt{0.05}.}\label{tab:hyperparams_det_coco}
        \resizebox{0.8\linewidth}{!}{
            \begin{tabular}{l@{\hspace*{5mm}} c@{\hspace*{5mm}}c@{\hspace*{5mm}}c}
                \toprule
                 Hyperparameters & MaskRCNN \\
                \midrule
                Stochastic depth rate & [0.0, ..., 0.3]  \\
                Data augmentation & Multi scale RRC  \\
                Train epochs & 12 \\
                Batch size & 16 \\
                Optimizer & AdamW \\
                Peak learning rate & [5e-4, 2e-4, 1e-4] \\
                LR. decay schedule type & Step-wise \\
                LR. decay schedule & [8, 11] \\
                Weight decay rate & 0.1 \\
                
                \bottomrule
        \end{tabular}
        }
\end{minipage}%
\hspace{10px}
\begin{minipage}[t!]{0.47\textwidth}%
        \centering
        \captionof{table}{Training hyperparameters for semantic segmentation experiments on ADE20k. ``RRC'' is \texttt{RandomResizedCrop}. We sweep through stochastic depth rate in steps of \texttt{0.05}.}\label{tab:hyperparams_segm}
        \resizebox{0.85\linewidth}{!}{
            \begin{tabular}{l@{\hspace*{5mm}} c@{\hspace*{5mm}}c@{\hspace*{5mm}}c}
                \toprule
             Hyperparameters & UperNet & SemanticFPN \\
            \midrule
            Stochastic depth rate & \multicolumn{2}{c}{[0.0, ..., 0.2]}  \\
            Data augmentation & \multicolumn{2}{c}{RRC}  \\
            Crop Size & \multicolumn{2}{c}{512$\times$512}  \\
            Train iterations & 160k & 40k \\
            Batch size & 16 & 64 \\
            Optimizer & \multicolumn{2}{c}{AdamW} \\
            Peak learning rate & \multicolumn{2}{c}{[5e-4, 2e-4, 1e-4]} \\
            LR. decay schedule type & \multicolumn{2}{c}{Polynomial} \\
            Warmup iterations & 1500 & - \\
            Weight decay rate & 0.01 & 5e-4 \\
            
            \bottomrule
        \end{tabular}
        }
\end{minipage}%

\begin{table}[!htb]
\centering
\caption{Training hyperparameters for depth estimation experiments on NYUv2 dataset. ``RRC'' is \texttt{RandomResizedCrop}. We sweep through stochastic
depth rate in steps of \texttt{0.05}.}
\scalebox{0.85}{
\begin{tabular}{l|c}
\toprule
 Hyperparameters & Value \\
\midrule
Stochastic depth rate & [0.0, ..., 0.2]  \\
Data augmentation &RRC  \\
Crop Size & 480$\times$480  \\
Train epochs & 25 \\
Batch size & 24 \\
Optimizer & AdamW \\
Peak learning rate & [7e-4, 5e-4, 2e-4, 1e-4] \\
Layer decay rate & 0.8 \\
Weight decay rate & 0.05 \\

\bottomrule

\end{tabular}
}
\label{tab:hyperparams_depth}
\end{table}

\end{document}